\documentclass[nohyperref]{article}

\usepackage{times}
\usepackage{latexsym}
\usepackage{xspace}
\usepackage{tabularx}

\usepackage{graphicx}
\usepackage{caption}
\usepackage{subcaption}
\usepackage{tikz}
\usepackage{pgfplots}
\usepackage{ pgfplotstable}
\usepgfplotslibrary{fillbetween}
\usetikzlibrary{patterns}
\pgfplotsset{compat=1.8}

\newenvironment{customlegend}[1][]{%
    \begingroup
    \csname pgfplots@init@cleared@structures\endcsname
    \pgfplotsset{#1}%
}{%
    \csname pgfplots@createlegend\endcsname
    \endgroup
}%

\def\addlegendimage{\csname pgfplots@addlegendimage\endcsname}

\usepackage{microtype}
\usepackage{booktabs} 

\usepackage{hyperref}

\usepackage[accepted]{icml2023}

\usepackage{amsmath}
\usepackage{amssymb}
\usepackage{mathtools}
\usepackage{amsthm}
\usepackage{nicefrac}

\usepackage[capitalize,noabbrev]{cleveref}

\theoremstyle{plain}

\theoremstyle{definition}

\theoremstyle{remark}

\usepackage[textsize=tiny]{todonotes}

\newcommand{\modelname}{\textsc{lumen}\xspace}
\newcommand{\memoryname}{MemoryFiD\xspace}
\newcommand{\papertitle}{Pre-computed memory or on-the-fly encoding? A hybrid approach to retrieval augmentation makes the most of your compute}

\newcommand{\modelnameonethird}{\modelname$\nicefrac{1}{3}$\xspace}
\newcommand{\modelnameoneeighth}{\modelname$\nicefrac{1}{8}$\xspace}

\newcommand{\corpussize}{\text{Corpus size}}
\newcommand{\liveprop}{\alpha}

\definecolor{modelcolor}{rgb}{0, 0.46484375, 0.73046875} 
\definecolor{fidcolor}{rgb}{0.9296875, 0.3984375, 0.46484375} 
\definecolor{memorycolor}{rgb}{0.95, 0.52, 0.0} 
\definecolor{finetunecolor}{rgb}{0.98, 0.85, 0.37} 
	
\definecolor{conditioncolor}{rgb}{0.18, 0.55, 0.34} 
\definecolor{q2memcolor}{rgb}{1.0, 0.57, 0.64}
\definecolor{mem2qcolor}{rgb}{0.66, 0.89, 0.63}
\definecolor{frozenqcolor}{rgb}{0.88, 0.66, 0.37}
\definecolor{decodercolor}{rgb}{0.9296875, 0.3984375, 0.46484375} 
\definecolor{encodercolor}{rgb}{0.265625, 0.46484375, 0.6640625} 
\definecolor{darkgrey}{rgb}{0.33203125,0.33203125,0.33203125} 
\definecolor{finetunecompcolor}{rgb}{0, 0.46484375, 0.73046875} 
\definecolor{frozencompcolor}{rgb}{0.95, 0.52, 0.0} 

\definecolor{largewidthcolor}{rgb}{0.85, 0.75, 0.85} 
\definecolor{basewidthcolor}{rgb}{1.0, 0.7, 0.28} 
\definecolor{smallwidthcolor}{rgb}{0.21, 0.46, 0.53} 

\definecolor{transferbasecolor}{rgb}{0.85, 0.75, 0.85} 
\definecolor{transfermemorycolor}{rgb}{1.0, 0.7, 0.28} 
\definecolor{transferlivecolor}{rgb}{0.21, 0.46, 0.53} 

\newcommand{\modelmark}{square*}
\newcommand{\fidmark}{diamond*}
\newcommand{\memorymark}{triangle*}
\newcommand{\finetunemark}{*}
\newcommand{\conditionmark}{triangle*}
\newcommand{\basemark}{diamond*}
\newcommand{\largemark}{triangle*}
\newcommand{\xlmark}{*}
\newcommand{\xxlmark}{square*}

\definecolor{basecolor}{rgb}{0.85, 0.75, 0.85} 
\definecolor{largecolor}{rgb}{1.0, 0.7, 0.28} 
\definecolor{xlcolor}{rgb}{0.05, 0.5, 0.06} 
\definecolor{xxlcolor}{rgb}{0, 0.46484375, 0.73046875} 

\definecolor{gapcolor13}{rgb}{0, 0.46484375, 0.73046875} 
\definecolor{gapcolor18}{rgb}{0.95, 0.52, 0.0} 
\newcommand{\propaxis}{Live proportion $\liveprop$}

\icmltitlerunning{\papertitle}

\begin{document}

\twocolumn[
\icmltitle{\papertitle}



\icmlsetsymbol{equal}{*}

\begin{icmlauthorlist}
\icmlauthor{Michiel de Jong}{equal,usc}
\icmlauthor{Yury Zemlyanskiy}{equal,google}
\icmlauthor{Nicholas FitzGerald}{google}
\icmlauthor{Joshua Ainslie}{google}\\
\icmlauthor{Sumit Sanghai}{google}
\icmlauthor{Fei Sha}{google}
\icmlauthor{William W. Cohen}{google}
\end{icmlauthorlist}

\icmlaffiliation{usc}{University of Southern California}
\icmlaffiliation{google}{Google Research}

\icmlcorrespondingauthor{Michiel de Jong}{msdejong@usc.edu}
\icmlcorrespondingauthor{Yury Zemlyanskiy}{urikz@google.com}

\icmlkeywords{Machine Learning, ICML}

\vskip 0.3in
]



\printAffiliationsAndNotice{\icmlEqualContribution} 

\begin{abstract}
Retrieval-augmented language models such as Fusion-in-Decoder are powerful, setting the state of the art on a variety of knowledge-intensive tasks. However, they are also expensive, due to the need to encode a large number of retrieved passages. Some work avoids this cost by pre-encoding a text corpus into a memory and retrieving dense representations directly. However, pre-encoding memory incurs a severe quality penalty as the memory representations are not conditioned on the current input. We propose \modelname, a hybrid between these two extremes, pre-computing the majority of the retrieval representation and completing the encoding on the fly using a live encoder that is conditioned on the question and fine-tuned for the task. We show that \modelname significantly outperforms pure memory on multiple question-answering tasks while being much cheaper than FiD, and outperforms both for any given compute budget. Moreover, the advantage of \modelname over FiD increases with model size.
\end{abstract}
    
\section{Introduction}

Retrieval-augmented language models such as Fusion-in-Decoder \citep{fid} achieve strong performance on knowledge intensive tasks, often outperforming much larger models \citep{atlas}. They retrieve related text passages and process the passages along with the input to extract relevant context information. However, encoding retrieved passages can be computationally expensive. Recent work has found that with an optimized decoder \citep{multiquery, fido, palminference} encoding retrieved passages makes up the bulk of total cost for finetuning and inference. 

\begin{figure}
     \centering
        \begin{tikzpicture}[scale=1.0]
            \begin{axis}[
            scale only axis,
            width=0.85\columnwidth,
            height=0.45\columnwidth,
            ylabel={Exact Match},
            xlabel={Proportion of live encoder layers},
            mark=x,
            ymajorgrids=true,
            xmajorgrids=true,
            xminorticks=true,
            grid style=dashed,
            legend columns=1,
            legend cell align=left,
            legend style={
                anchor=south,
                at={(0.8, 0.3)},
            },
        ]
            \addplot[color=memorycolor,line width=3, dotted] table {
                0 42.72  
            	1.0 42.72
            };
            \addplot[color=xxlcolor,mark=\xxlmark,mark size=2pt,line width=2] table {
                0 42.72  
                0.041666 44.26
                0.125 52.89
                0.25 54.41
                0.333 54.95
                0.5 55.0
                0.75 55.14
            	1.0 55.21
            };             
            \addplot[color=fidcolor,mark size=2pt,line width=3, dotted] table {
                0 55.21
            	1.0 55.21
            };
            \legend{Memory, \modelname, FiD}
            \end{axis}
        \end{tikzpicture}
    \caption{Exact match on Natural Questions dev set for \modelname-XXL as a function of proportion of live (fine-tuned and conditioned on question) vs memory (pre-computed) encoder layers. \modelname closes the gap between pure memory and FiD approaches with a fraction of live layers and therefore compute.}
    \label{fig:perf_prop_intro}
\end{figure}
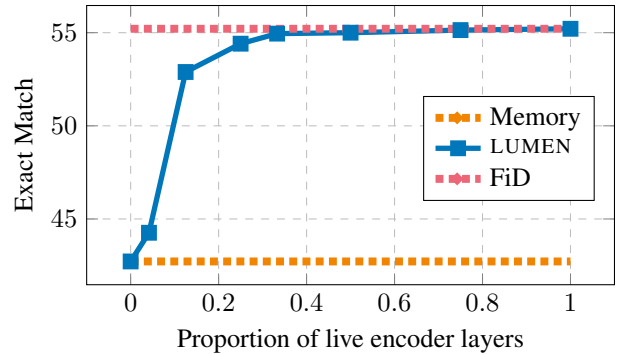
\begin{figure*}[t!]
  \centering
  \includegraphics[width=\linewidth]{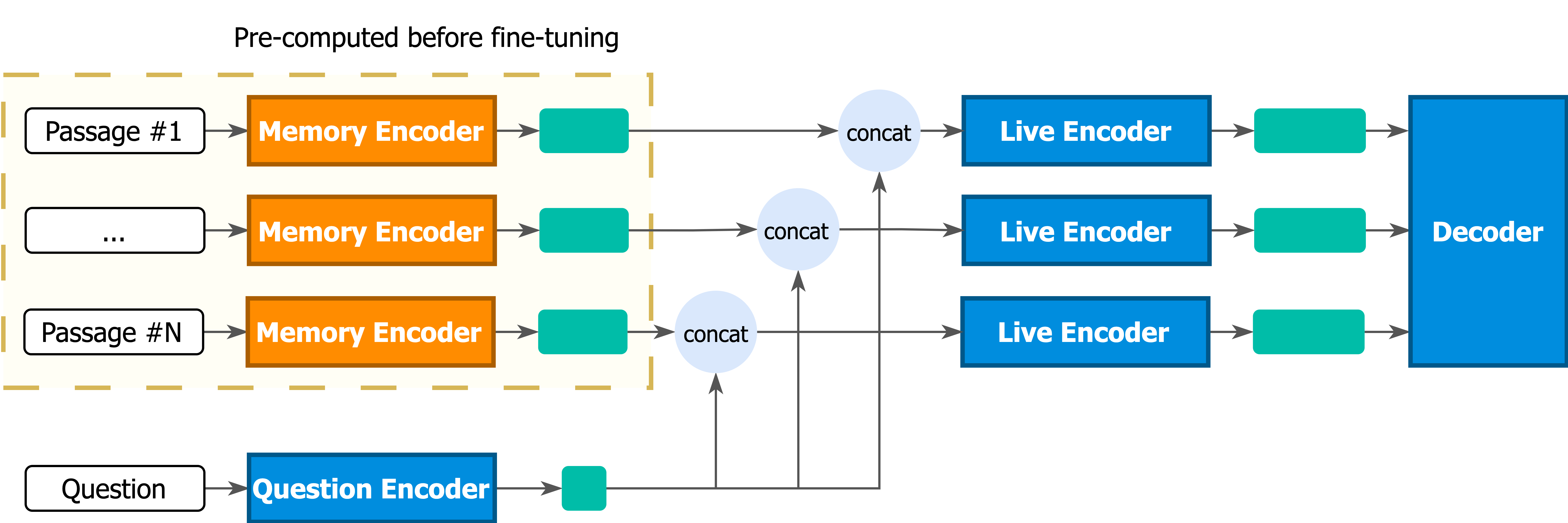}
  \caption{Overview of the \modelname architecture. Before fine-tuning, each passage in the corpus is encoded by a memory encoder. While processing a sample, a question encoder first generates a representation of the question, which is then separately concatenated with each pre-computed passage representation. A fine-tuned live encoder then updates the passage representations conditioning on the question, which are finally fed into the decoder as in standard FiD. Frozen components are in orange, fine-tuned components in blue.}
  \label{fig:architecture}
\end{figure*}

An increasingly common approach to reduce this encoding cost is to retrieve and extract information from a memory of pre-computed representations rather than raw text, amortizing the encoding of a passage over every sample that retrieves the passage entry from memory \citep{fidmemory, tome, memorizing, trime, qama, emat}.\footnote{Here we do not refer to pre-computing representations used to select passages for retrieval (as is common practice for dense retrieval methods), but rather pre-computing the actual representations to be retrieved and incorporated into the language model.}

However, memory approaches incur a large quality penalty relative to retrieval-augmented models~\citep{fidmemory}, because the pre-encoded memory is not conditioned on the task or on the particular input or question. Thus, the pre-encoded representation must be sufficiently comprehensive to answer \emph{any} question, a challenging undertaking. The human analogue is the difference between memorizing an entire book and being quizzed afterwards compared to looking up the answer to a question on the fly.

Memory-based approaches therefore need to massively scale model size in order to achieve comparable performance. As we will show, this leads to higher overall net FLOPs due to cross-attention and decoding, as well as impractical increases in pre-training, pre-computation, and storage costs. 

We propose \modelname (Live Update Memory Network), a middle ground between retrieval and memory. \modelname divides the task of encoding passages between a frozen memory encoder that pre-computes passage memory representations, and a fine-tuned live encoder that updates the memory representations conditioned on the question. Figure \ref{fig:architecture} provides a detailed overview of the architecture. As can be seen in Figure \ref{fig:perf_prop_intro}, a small proportion of live layers can already achieve performance close to standard Fusion-in-Decoder. 

We start with a set of experiments initializing \modelname from T5, partitioning the standard T5 encoder into a memory and live encoder. We evaluate on question-answering datasets Natural Questions~\citep{nq} and TriviaQA~\citep{triviaqa}. \modelname achieves significantly stronger performance than FiD and FiD with memory given the same computational budget, and the advantage increases with model size. At T5-XXL size \modelname performs comparably to FiD with only one third proportion of live layers and FLOPs.

Next, we experiment with improvements to the standard \modelname setup, showing that the performance-compute trade-off can be further improved relative to FiD by transferring a trained memory and live encoder from a related task. In particular, we find that transfer from Natural Questions can close most of the gap in performance between \modelname and FiD with smaller live encoders or on small datasets. 
Ultimately, \modelname represents a desirable trade-off between retrieval and memory-based approaches, achieving better performance for any given computational budget.
\section{Background}
\begin{figure*}
     \centering
     \begin{subfigure}[t]{0.48\textwidth}
        \raggedright
        \begin{tikzpicture}[scale=1.0]
            \begin{axis}[
            scale only axis,
            width=0.9\columnwidth,
            height=0.5\textwidth,            
            ylabel={Exact Match},
            xlabel={\propaxis},
            mark=x,
            ymajorgrids=true,
            xmajorgrids=true,
            xminorticks=true,
            grid style=dashed,
            legend columns=1,
            legend cell align=left,
            legend pos={south east},
            ]
            \addlegendimage{empty legend}\addlegendentry{NaturalQ}
            \addplot[color=basecolor,mark=\basemark,mark size=1pt,line width=2] table {
                0 34.3
                0.041666 36.69
                0.125 39.5
                0.25 42.91
                0.333 44.47
                0.5 46.91
                0.75 46.08
            	1.0 46.55
            };
            \addplot[color=largecolor,mark=\largemark,mark size=1pt,line width=2] table {
                0 38.34
                0.041666 39.75
                0.125 42.78
                0.25 48.76
                0.333 50.46
                0.5 51.2
                0.75 51.52
            	1.0 51.79
            };
            \addplot[color=xlcolor,mark=\xlmark,mark size=1pt,line width=2] table {
                0 40.29  
                0.041666 41.86
                0.125 46.73
                0.25 50.54
                0.333 53.09
                0.5 53.11
                0.75 53.24
            	1.0 53.81
            };
            \addplot[color=xxlcolor,mark=\xxlmark,mark size=1pt,line width=2] table {
                0 42.72  
                0.041666 44.26
                0.125 52.89
                0.25 54.41
                0.333 54.95
                0.5 55.0
                0.75 55.14
            	1.0 55.21
            };            
            \end{axis}
        \end{tikzpicture}  
     \end{subfigure} \hspace{0.36cm}
     \hfill
     \begin{subfigure}[t]{0.48\textwidth}
        \centering
        \begin{tikzpicture}[scale=1.0]
            \begin{axis}[
            scale only axis,
            width=0.9\columnwidth,
            height=0.5\textwidth,            
            xlabel={\propaxis},
            mark=x,
            ymajorgrids=true,
            xmajorgrids=true,
            xminorticks=true,
            grid style=dashed,
            legend columns=1,
            legend cell align=left,
            legend pos={south east},
            ]
            \addlegendimage{empty legend}\addlegendentry{TriviaQA}
            \addplot[color=basecolor,mark=\basemark,mark size=1pt,line width=2] table {
                0 50.18  
                0.041666 52.36
                0.125 55.35
                0.25 61.81
                0.333 62.53
                0.5 62.88
                0.75 64.05
            	1.0 63.28
            };
            \addplot[color=largecolor,mark=\largemark,mark size=1pt,line width=2] table {
                0 55.05  
                0.041666 57.3
                0.125 61.94
                0.25 65.96
                0.333 67.16
                0.5 67.84
                0.75 68.17
            	1.0 67.64
            };
            \addplot[color=xlcolor,mark=\xlmark,mark size=1pt,line width=2] table {
                0 59.36  
                0.041666 61.24
                0.125 66.62
                0.25 68.59
                0.333 69.83
                0.5 70.63
                0.75 70.52
            	1.0 71.09
            };
            \addplot[color=xxlcolor,mark=\xxlmark,mark size=1pt,line width=2] table {
                0 63.01
                0.041666 64.81
                0.125 71.03
                0.25 72.15
                0.333 72.58
                0.5 72.33
                0.75 72.35
            	1.0 72.8
            };         
            \end{axis}
        \end{tikzpicture}  
     \end{subfigure}     
     \hfill
     
     \begin{tikzpicture}
        \begin{customlegend}[
            legend columns=4,
            legend style={
                align=center,
                column sep=4ex,
                font=\large,
            },
            legend entries={Base, Large, XL, XXL}
        ]
            \addlegendimage{mark=\basemark,solid,color=basecolor,line width=2}
            \addlegendimage{mark=\largemark,mark size=3pt,solid,color=largecolor,line width=2}
            \addlegendimage{mark=\xlmark,mark size=3pt,solid,color=xlcolor,line width=2}
            \addlegendimage{mark=\xxlmark,mark size=3pt,solid,color=xxlcolor,line width=2}   
        \end{customlegend}
    \end{tikzpicture}            
     
    \caption{\textbf{MAIN RESULT: \modelname achieves performance close to FiD with fraction of live layers. The required fraction decreases with scale.} Exact match on Natural Questions (NaturalQ) and TriviaQA validation sets as a function of proportion of live encoder layers for \modelname Base, Large, XL, and XXL models.}
    \label{fig:perf_vs_liveprop}
\end{figure*}
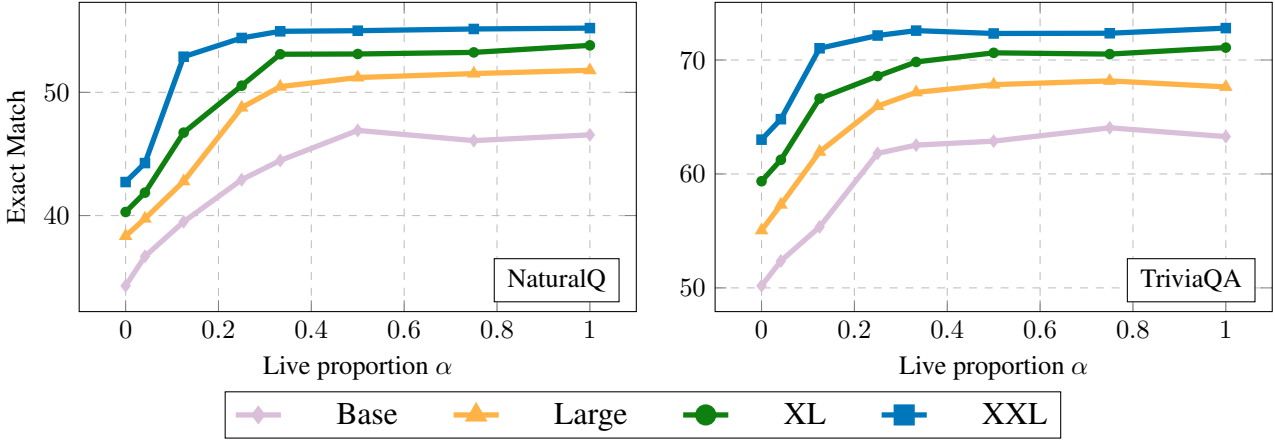

We are interested in achieving the best possible performance for any given resource budget. However, there are different types of computational resources, and varying algorithmic approaches yield distinct trade-offs between those resources. In this section we provide background on existing retrieval-augmented models and describe the costs of those models along different computational dimensions.

\subsection{Computational resources for retrieval-augmented models}

The usual life-cycle of current models starts with pre-training, followed by fine-tuning on multiple tasks. Finally, the model is used for inference, either online or for batch distillation to a smaller model. Each of these stages features a different cost per sample. Let $N_{\mathit{pt}}, N_{\mathit{ft}}$ and $N_{I}$ be the number of processed samples for pre-training, fine-tuning and inference, and $F_{\mathit{pt}}, F_{\mathit{ft}}$ and $F_I$ the compute cost per sample for each stage, measured in FLOPs (floating point operations). Then the compute costs for the model are
\begin{align*}
    &\text{FLOPs}_{\text{pre-train}} = N_{\mathit{pt}} F_{\mathit{pt}} \\
    &\text{FLOPs}_{\text{fine-tune}} = N_{\mathit{ft}} F_{\mathit{ft}} \cdot \text{number of tasks}\\
    &\text{FLOPs}_{\text{inference}} = N_I F_I
\end{align*}
The methods proposed in this paper are agnostic to the method used to select retrieved passages, so we do not consider retrieval method in our comparison of computational cost. While FiD inference can be slower than what FLOPs would indicate due to decoder memory bandwidth constraints \citep{fidlight, fido}, modifying the decoder mitigates the gap~ \citep{fido}. Hence, we use FLOPs our measure of computation cost in line with related work \cite{kgfid,canext}. Appendix \ref{section:flops_vs_latency} contains additional evidence on the relationship between FLOPs and latency.

For retrieval-augmented models there are additional costs. The retrieval set must be stored and retrievals transmitted to the accelerator. There may also be preprocessing overhead for the retrieval set, such as pre-computing memory representations. Let $N_{rs}$ be the size of the retrieval set
and $F_{\mathit{pc}}$ the FLOPs associated with preprocessing a retrieval candidate. Then storage requirements and pre-computaton costs are given by
\begin{align*}
    \text{Storage} &= \corpussize \cdot \text{Size of a single sample}\\
    \text{FLOPs}_{\text{precompute}} &= \corpussize \cdot F_{\text{precompute}}
\end{align*}
If retrieval representations are fine-tuned, then a different version of the retrieval set must be pre-computed and stored for each task. Required bandwidth for transmission is determined by the product of the number and size of retrieved representations.

\subsection{Fusion-in-Decoder}
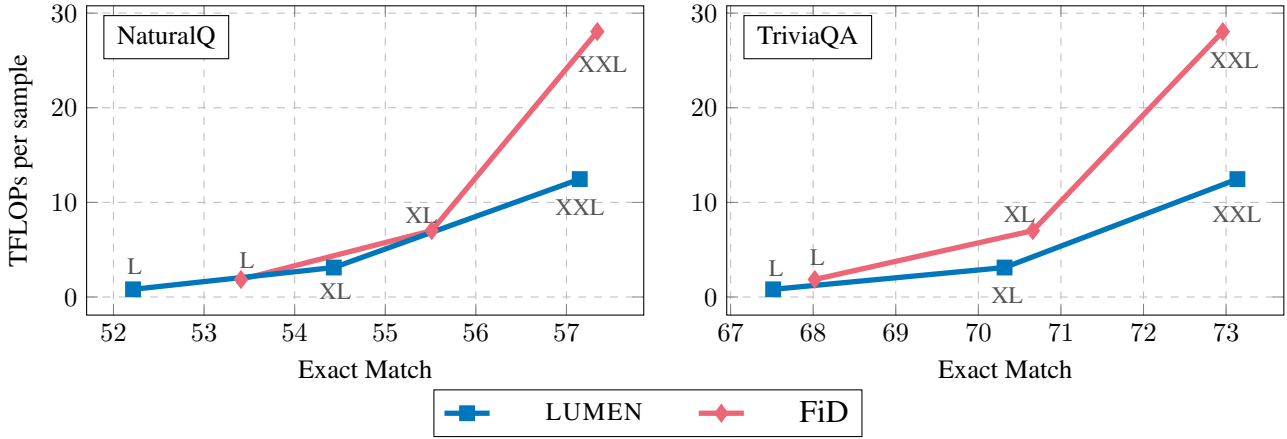
\begin{figure*}
     \centering
     \begin{subfigure}[t]{0.48\textwidth}
        \raggedright
                \begin{tikzpicture}[scale=1.0]
            \begin{axis}[
            scale only axis,
            width=0.9\textwidth,
            height=0.5\textwidth,
            xlabel={Exact Match},
            ylabel={TFLOPs per sample},
            mark=x,
            x tick label style={log ticks with fixed point},
            ymajorgrids=true,
            xmajorgrids=true,
            xminorticks=true,
            grid style=dashed,
            legend pos={north west},
            ylabel style={align=left, text height=0.2cm},
            ]
            \addlegendimage{empty legend}\addlegendentry{NaturalQ}
            \addplot[color=fidcolor,mark=\fidmark,mark size=2pt,line width=2] table {
                53.40720222 1.850443163
                55.51246537 7.012205323
                57.34072022 28.04882092 
            };
            \addplot[color=modelcolor,mark=\modelmark,mark size=2pt,line width=2] table {
                52.21606648199446 0.8125205852 
            	54.43213296 3.115193481 
            	57.1468144 12.46077388         	
            };
            
            \node at (axis cs:53.29,5.68) [anchor= north west, color=darkgrey] {\small{L}};
            \node at (axis cs:55.4,10.5) [anchor= north, color=darkgrey] {\small{XL}};
            \node at (axis cs:57.4,26.4) [anchor= north, color=darkgrey] {\small{XXL}};
            
            \node at (axis cs:52.05,5.1) [anchor= north west, color=darkgrey] {\small{L}};
            \node at (axis cs:54.45,2.5) [anchor= north, color=darkgrey] {\small{XL}};
            \node at (axis cs:57.15,11.3) [anchor= north, color=darkgrey] {\small{XXL}};            
            
            \end{axis}
        \end{tikzpicture}  
     \end{subfigure} \hspace{0.36cm}
     \hfill
     \begin{subfigure}[t]{0.48\textwidth}
        \centering
                \begin{tikzpicture}[scale=1.0]
            \begin{axis}[
            scale only axis,
            width=0.9\textwidth,
            height=0.5\textwidth,
            xlabel={Exact Match},
            mark=x,
            x tick label style={log ticks with fixed point},
            ymajorgrids=true,
            xmajorgrids=true,
            xminorticks=true,
            grid style=dashed,
            legend pos={north west},
            ylabel style={align=left, text height=0.2cm},
            ]
            \addlegendimage{empty legend}\addlegendentry{TriviaQA}
            \addplot[color=fidcolor,mark=\fidmark,mark size=2pt,line width=2] table {
                68.01909308 1.850443163
                70.66 7.012205323
                72.96031115 28.04882092  
            };
            \addplot[color=modelcolor,mark=\modelmark,mark size=2pt,line width=2] table {
                67.51524794 0.8125205852 
            	70.31733404 3.115193481
            	73.13709891 12.46077388        	
            };
                
            \node at (axis cs:67.85,6.05) [anchor= north west, color=darkgrey] {\small{L}};
            \node at (axis cs:70.5,10.6) [anchor= north, color=darkgrey] {\small{XL}};
            \node at (axis cs:73.1,26.9) [anchor= north, color=darkgrey] {\small{XXL}};
            
            \node at (axis cs:67.35,4.9) [anchor= north west, color=darkgrey] {\small{L}};
            \node at (axis cs:70.35,1.99) [anchor= north, color=darkgrey] {\small{XL}};
            \node at (axis cs:73.13,10.6) [anchor= north, color=darkgrey] {\small{XXL}};       
            \end{axis}
        \end{tikzpicture}  
     \end{subfigure}     
     \hfill
     \begin{tikzpicture}
        \begin{customlegend}[
            legend columns=2,
            legend style={
                align=center,
                column sep=4ex,
                font=\large,
            },
            legend entries={\modelname,FiD}
        ]
        \addlegendimage{mark=\modelmark,mark size=2pt,solid,color=modelcolor,line width=2}
        \addlegendimage{mark=\fidmark,solid,color=fidcolor,line width=2,mark size=2pt}
        \end{customlegend}
    \end{tikzpicture}            
     
    \caption{\textbf{MAIN RESULT: \modelname uses significantly less compute than FiD for the same performance, and this advantage grows with scale.} TFLOPs as a function of exact match on Natural Questions (NaturalQ) and TriviaQA test sets. FLOPs are for single forward step and exclude pre-computation. Compares FiD and \modelname with live proportion 0.33 Large, XL and XXL models. Lower is better.}
    \label{fig:perf_vs_time}
\end{figure*}

Fusion-in-Decoder \citep{fid} consists of a T5 encoder-decoder model. For each input, a number of relevant text passages are retrieved, and the input is prepended to each passage. The resulting passages are encoded separately by the encoder, and the encoded representations are then concatenated and attended to by the decoder to produce a target output. For each model, \textcolor{finetunecompcolor}{\textbf{fine-tuned}} components are in blue and \textcolor{frozencompcolor}{\textbf{frozen}} components in orange.
\begin{equation*}
    G = \text{\textbf{\textcolor{finetunecompcolor}{Dec}}}\Big[\text{\textcolor{finetunecompcolor}{\textbf{Enc}}}(Q; \text{Passage}_1); \ldots \text{\textbf{\textcolor{finetunecompcolor}{Enc}}}(Q; \text{Passage}_k)\Big]
\end{equation*}
Let $n_s$ be the number of source tokens, $n_t$ the number of target tokens, $L$ the number of layers, and $d$ the dimension of the model. Following the analysis from \citet{fido}, the FLOPs for a single inference sample of FiD (ignoring attention score computation) is given by\footnote{We approximate the FLOPS of the MLP block as $8d^2$, the FLOPs from the original Transformer MLP. The T5 MLP has dimension between $2.5d$ and $3d$ and three matrix multiplication operations including GEGLU, yielding total FLOPs close to 8d.}
\begin{equation*}
    F_{I} = \underbrace{n_s \cdot L \cdot 14 d^2}_{\text{Encoder and cross-attention}} + \underbrace{n_t \cdot L \cdot 14 d^2}_{\text{Decoder}}
\end{equation*}
Appendix \ref{section:complexity} discusses the derivation of this complexity in greater detail. $F_{pt}$ and $F_{ft}$ equal $3 F_{I}$ due to the backward step. For fine-tuning and inference $n_s \gg n_t$ because of the large number of tokens from the retrieved passages. As such, FiD's fine-tuning and inference FLOPs per sample are significantly higher than for pre-training. In contrast, storage and bandwidth requirements are low as the retrieval set consists of passages of raw tokens. FiD has no pre-computation costs.

\subsection{Memory}

An increasing number of works reduce the cost of retrieval-augmented models by pre-computing dense representations of retrieval candidates and storing them in a memory. One such work modifies FiD by pre-computing passage encoder representations and providing the input as a prefix to the decoder \citep{fidmemory}. We refer it  as \memoryname: 
\begin{equation*}
    G = \text{\textbf{\textcolor{finetunecompcolor}{Dec}}}\Big[Q; \text{\textbf{\textcolor{frozencompcolor}{MemEnc}}}(\text{Passage}_1);.. \text{\textbf{\textcolor{frozencompcolor}{MemEnc}}}(\text{Passage}_k)\Big]
\end{equation*}

\memoryname saves fine-tuning and inference compute at the expense of increased pre-computation, storage, and bandwidth requirements. Because \memoryname does not encode retrieved passages on the fly, encoder costs are removed and only cross-attention and other decoder compute are left:
\begin{equation*}
    F_{I} = \underbrace{n_s \cdot L \cdot 2 d^2}_{\text{Cross-attention}} + \underbrace{n_t \cdot L \cdot 14 d^2}_{\text{Decoder}}
\end{equation*}
Instead, it pre-computes passage representations, using
\begin{equation*}
    \text{FLOPs}_{\text{precompute}} = \corpussize \cdot n_p L \cdot 12 d^2
\end{equation*}
where $n_p$ is the number of tokens in a single passage. \memoryname stores the final layer representations for each passage token, taking up
\begin{equation*}
    \text{Storage} = \corpussize \cdot 2n_p d
\end{equation*}
To give an indication of the storage overhead, applying \memoryname-XXL to Wikipedia requires approximately 16 terabytes. Holding model size fixed, \memoryname saves compute as long as the retrieval corpus is not too large relative to the number of samples processed for fine-tuning and inference. However, as passage representations are not conditioned on the question, \memoryname incurs a significant performance penalty relative to normal FiD. Therefore, in order to reach equivalent performance to standard FiD, \memoryname must use a much larger model, which incurs much larger cross-attention, decoder, pre-training, pre-computation, storage and bandwidth costs. \citet{fidmemory} also fine-tune the memory encoder, which requires pre-computing and storing a separate memory for each task. This is intractable for real applications involving internet-sized corpora, so for our main results we assume the memory is pre-computed from a single model without fine-tuning on individual tasks.  Without fine-tuning, the performance penalty is even higher. Figure \ref{fig:perf_vs_prop_ablation} shows the effect of fine-tuning memory; \modelname results are qualitatively similar in that setting.

\begin{table}[h!]
\centering
\caption{Model FLOPs per layer and storage bytes per token for FiD, MemoryFiD and \modelname.}
\vspace{0.35cm}
\begin{tabular}{lcc}
\toprule
\textbf{Model} & \textbf{FLOPs} & \textbf{Storage} \\
\midrule
\textsc{FiD}& $14 n_s d^2 + 14 n_t d^2$ & 10 \\
\textsc{MemFiD}& $2 n_s d^2 + 14 n_t d^2$ & 2d \\
\textsc{\modelname}& $(2 + 12\liveprop) n_s d^2 + 14 n_t d^2$ & 2d \\
\bottomrule
\end{tabular}
\label{table:complexity_storage}
\end{table}
\section{\modelname}

Intuitively when reading a passage it is helpful to know what information is needed and for what purpose. For Fusion-in-Decoder, this is achieved by prepending the input to retrieved passages and fine-tuning the passage encoder, whereas \memoryname does not enjoy such an advantage. With \modelname, we explore the possibility that a similar effect can be achieved by a two-step process, in which a large model generates a general representation for each passage that can be placed in memory, and a smaller model transforms this general representation into an input-specific representation by conditioning on the input and task. Figure \ref{fig:architecture} provides an overview of the \modelname architecture.

\subsection{Architecture}
\begin{figure*}
     \centering
     \begin{subfigure}[t]{0.48\textwidth}
        \raggedright

        \begin{tikzpicture}[scale=1.0]
            \begin{axis}[
            scale only axis,
            width=0.9\textwidth,
            height=0.5\textwidth,
            ylabel={Exact Match},
            xlabel={TFLOPs per sample},
            mark=x,
            x tick label style={log ticks with fixed point},
            ymajorgrids=true,
            xmajorgrids=true,
            xminorticks=true,
            grid style=dashed,
            legend pos={north east},
            ylabel style={align=left, text height=0.2cm},
            ]
            \addlegendimage{empty legend}\addlegendentry{NaturalQ}
            \addplot[color=memorycolor,mark=\memorymark,mark size=2pt,line width=2] table {
                0.2808505958 43.29639889196676
                1.119375852 47.20221607
                4.477503406 49.1966759
            };
            \addplot[color=modelcolor,mark=\modelmark,mark size=2pt,line width=2] table {
                0.1916377569 44.54293629
                0.6888725033 52.21606648199446
            	2.665564092  54.43213296
            
            };
            \node at (axis cs:0.28,43.6) [anchor= north west, color=darkgrey] {\small{L}};
            \node at (axis cs:1.15,47.0) [anchor= north, color=darkgrey] {\small{XL}};
            \node at (axis cs:4.5,48.9) [anchor= north, color=darkgrey] {\small{XXL}};
            
            \node at (axis cs:-0.1,46.2) [anchor= north west, color=darkgrey] {\small{B}};
            \node at (axis cs:0.78,52) [anchor= north, color=darkgrey] {\small{L}};
            \node at (axis cs:2.69,54.0) [anchor= north, color=darkgrey] {\small{XL}};  
            \end{axis}
        \end{tikzpicture}  
     \end{subfigure} \hspace{0.36cm}
     \hfill
     \begin{subfigure}[t]{0.48\textwidth}
        \centering
      \begin{tikzpicture}[scale=1.0]
            \begin{axis}[
            scale only axis,
            width=0.9\textwidth,
            height=0.5\textwidth,
            xlabel={TFLOPs per sample},
            mark=x,
            x tick label style={log ticks with fixed point},
            ymajorgrids=true,
            xmajorgrids=true,
            xminorticks=true,
            grid style=dashed,
            legend pos={north east},
            ylabel style={align=left, text height=0.2cm},
            ]
            \addlegendimage{empty legend}\addlegendentry{TriviaQA}
            \addplot[color=memorycolor,mark=\memorymark,mark size=2pt,line width=2] table {
                0.2863870771 59.95757094
                1.139508511 62.41492089
                4.558034043 66.10094581
            };
            \addplot[color=modelcolor,mark=\modelmark,mark size=2pt,line width=2] table {
                0.2253599611 63.17510828
                0.8125205852 67.51524794 
            	3.115193481  70.31733404
            };
            \node at (axis cs:0.28,60.2) [anchor= north west, color=darkgrey] {\small{L}};
            \node at (axis cs:1.18,62.22) [anchor= north, color=darkgrey] {\small{XL}};
            \node at (axis cs:4.58,65.69) [anchor= north, color=darkgrey] {\small{XXL}};
            
            \node at (axis cs:0.063,62.85) [anchor= north west, color=darkgrey] {\small{B}};
            \node at (axis cs:0.86,67.2) [anchor= north, color=darkgrey] {\small{L}};
            \node at (axis cs:3.12,70.0) [anchor= north, color=darkgrey] {\small{XL}};  
            \end{axis}
        \end{tikzpicture}  
     \end{subfigure}     
     \hfill
     \begin{tikzpicture}
        \begin{customlegend}[
            legend columns=2,
            legend style={
                align=center,
                column sep=4ex,
                font=\large,
            },
            legend entries={\modelname, \memoryname}
        ]
        \addlegendimage{mark=\modelmark,mark size=2pt,solid,color=modelcolor,line width=2}          
        \addlegendimage{mark=\memorymark,solid,color=memorycolor,line width=2}
        \end{customlegend}
    \end{tikzpicture}            
     
    \caption{\textbf{\modelname achieves much better performance than \memoryname at any compute budget.} Exact match performance on the test set of Natural Questions as a function of TFLOPs per sample comparing \modelname~$\nicefrac13$ Base, Large and XL models with \memoryname Large, XL, and XXL models. FLOPs are for single forward step and exclude pre-computation. Note that axes are transposed relative to Figure \ref{fig:perf_vs_time} as \memoryname requires too much compute to match \modelname performance.}
    \label{fig:perf_vs_time_mem}
\end{figure*}
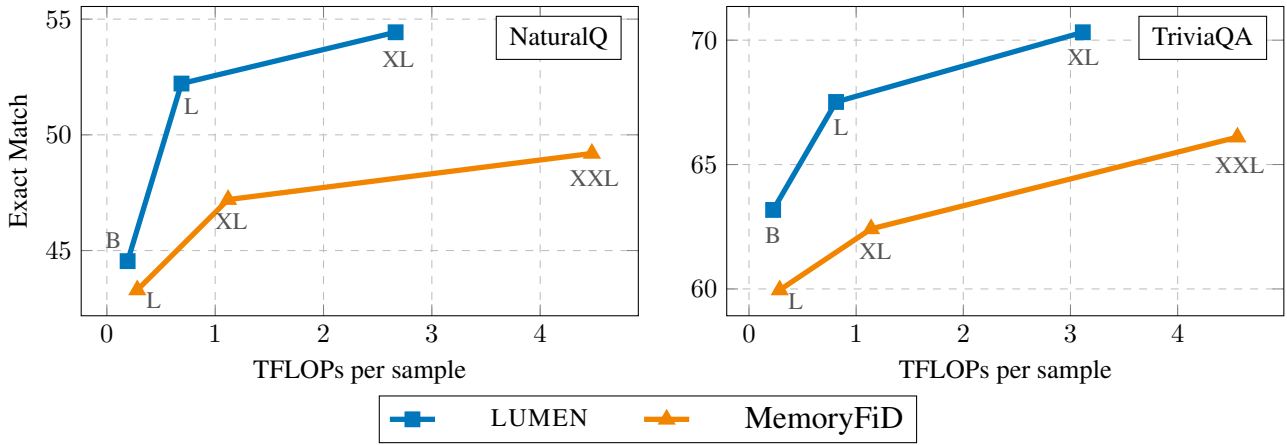

\modelname is initialized from a pre-trained T5 encoder-decoder model. The decoder functions the same as the standard FiD decoder, but \modelname features three encoders. The T5 encoder is divided into a large \textcolor{frozencompcolor}{\textbf{memory encoder}} which contains the first $1 - \liveprop$ proportion of layers, and a smaller \textcolor{finetunecompcolor}{\textbf{live encoder}} with the remaining $\liveprop$ proportion of layers. The memory encoder is applied offline to passages in the corpus to pre-compute memory representations, which are later updated conditioned on input and task on the fly by the fine-tuned live encoder. In order to ensure that memory representations and input are compatible, \modelname applies a \textcolor{finetunecompcolor}{\textbf{question encoder}} to the input before prepending the question representation to the memory representation. The question encoder shares its structure and initial weights with the memory encoder, but is fine-tuned.
\vspace{-3pt}
\begin{align*}
\label{eqn:hybrid}
    &G = \text{\textbf{\textcolor{finetunecompcolor}{Dec}}}\Big[Q; \text{\textbf{\textcolor{finetunecompcolor}{LiveEnc}}}(H_1); \ldots \text{\textbf{\textcolor{finetunecompcolor}{LiveEnc}}}(H_k) \Big] \\
    & H_i = \Big[\textbf{\textcolor{finetunecompcolor}{QEnc}}(Q);\hspace{0.2cm} \text{\textbf{\textcolor{frozencompcolor}{MemEnc}}}(\text{Passage}_i)\Big]
\end{align*} 
Choosing $\liveprop = 0$ recovers \memoryname, while $\liveprop = 1$ yields a model very close to FiD.
\vspace{-5pt}

\subsection{Computational analysis}
During fine-tuning and inference \modelname applies only a proportion $\liveprop$ of the layers, leading to a fraction $\liveprop$ of FiD reader FLOPs for any given model size.
\begin{equation*}
    F_{I} = \underbrace{n_s \cdot \liveprop L \cdot 12d^2}_{\text{Encoder}} + \underbrace{n_s \cdot L \cdot 2d^2}_{\text{Cross-attention}} + \underbrace{n_t \cdot L \cdot 14 d^2}_{\text{Decoder}}
\end{equation*}
Pre-computation costs at the same model size are a factor $1 - \liveprop$ of \memoryname pre-computation costs (\textit{without} fine-tuning the memory encoder). Storage and bandwidth costs are the same as for \memoryname (at same model size and without fine-tuning the memory encoder). Table \ref{table:complexity_storage} shows a comparison of FLOPs and storage requirements of FiD, MemoryFiD and \modelname. As we will show, \modelname can match FiD performance with only a modest increase in size, leading to a large decrease in computational cost without the commensurate increases in pre-training, pre-computation, and storage requirements incurred with \memoryname.

\section{Experiments}

\subsection{Experiment Setup}

\paragraph{Training procedure}

All experiments use models based on the T5.1.1 architecture \citep{t5}. The main experiments use models initialized from the public T5 checkpoints~\citep{t5check}. FiD is trained according to the standard recipe~\citep{fid}. For \modelname, given proportion of live layers $\liveprop$, the memory encoder and question encoder are each initialized with the first 1 - $\liveprop$ proportion of layers of the T5 encoder, and the live encoder is initialized with the last $\liveprop$ proportion of layers of the T5 encoder.

Models are fine-tuned with the T5X framework \citep{t5x} based on JAX \citep{jax} and FLAX \citep{flax} using the Adafactor \citep{adafactor} optimizer with batch size 64 and learning rate 0.0001. Test results are generated from checkpoints with the best dev results. Experiments in Section \ref{section:mem_shape} pre-train models from scratch. Pre-training follows the standard T5 training recipe except that we train for 500k steps, and disable the Adafactor second moment update schedule and factoring.

\paragraph{Data}

We evaluate \modelname on open-domain question-answering datasets Natural Questions \citep{nq}, TriviaQA \citep{triviaqa}, and WebQuestions~\citep{webquestions} (in Section \ref{section:transfer}). For all datasets, each sample is paired with the 20 most relevant 100-word Wikipedia passages ranked by DPR \citep{dpr} score. For FiD, the concatenated question and passage pairs are truncated to 256 tokens. For \modelname, the question and passage are individually truncated to 48 and 208 tokens to provide a fair comparison, as they are processed separately.
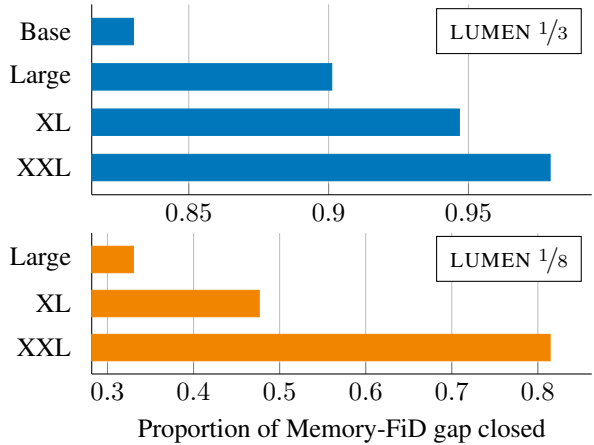
\begin{figure}[h!]
\begin{subfigure}[h!]{0.48\textwidth}
\begin{tikzpicture}[scale=1.0]
\begin{axis}[
    xbar,
    enlarge y limits=0.2,
    width=\columnwidth,
    height=0.5\columnwidth,
    major y tick style = transparent,    
    xmajorgrids = true,
    symbolic y coords={Base, Large, XL, XXL},
    ytick = data,
    y dir=reverse,
    axis y line*=none,
    axis x line*=bottom,
]
    \addlegendimage{empty legend}\addlegendentry{\modelnameonethird}
    \addplot[style={gapcolor13,fill=gapcolor13,mark=none}]
        coordinates {(0.8302040816,Base) (0.9011152416,Large) (0.9467455621,XL) (0.9791833467,XXL)};
\end{axis}
\end{tikzpicture}            
\end{subfigure}
\begin{subfigure}[h!]{0.48\textwidth}
\begin{tikzpicture}[scale=1.0]
\begin{axis}[
    xbar,
    enlarge y limits=0.3,
    width=\columnwidth,
    height=0.42\columnwidth,
    major y tick style = transparent,    
    xmajorgrids = true,
    xlabel = {Proportion of Memory-FiD gap closed},
    symbolic y coords={Large, XL, XXL},
    ytick = data,
    y dir=reverse,
    axis y line*=none,
    axis x line*=bottom,
]
    \addlegendimage{empty legend}\addlegendentry{\modelnameoneeighth}
    \addplot[style={gapcolor18,fill=gapcolor18,mark=none}]
        coordinates {(0.3301115242,Large) (0.4763313609,XL) (0.8142514011,XXL)};
\end{axis}
\end{tikzpicture}            
\end{subfigure}
\caption{\textbf{\modelname closes the gap with FiD as scale increases.} Proportion of exact match difference on Natural Questions between \memoryname and FiD closed by \modelname as a function of model scale.}
\label{fig:gap_scale}
\end{figure}
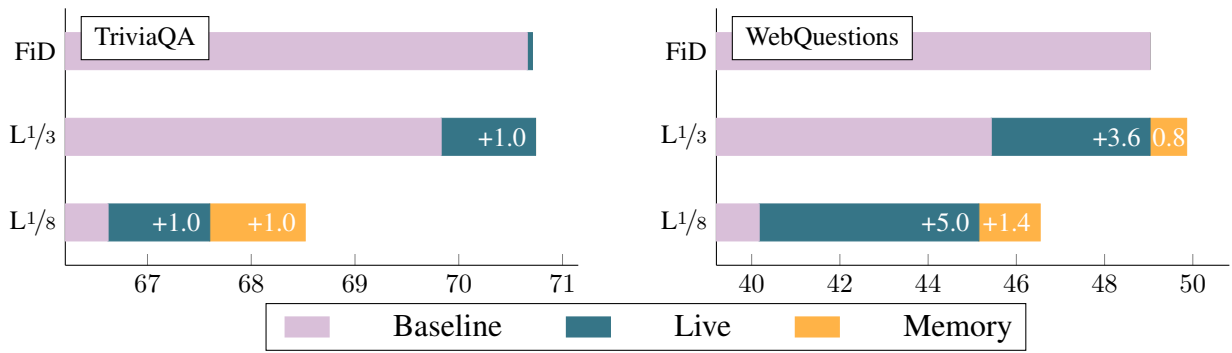
\begin{figure*}
\centering
\begin{subfigure}[t]{0.49\textwidth}
    \pgfplotstableread{ 
Label    Baseline Live Memory
FiD   70.66 0.05  0.0
L$\nicefrac{1}{3}$ 69.83    0.91  0.0
L$\nicefrac{1}{8}$ 66.62     0.98 0.92
    }\tqatransferdata

    \begin{tikzpicture}
    \begin{axis}[
    xbar stacked,   
    ytick=data,     
    y dir=reverse,
    enlarge y limits=0.25,
    bar width=14pt,
    height = 5cm,
    width = \textwidth,
    legend pos={north west},
    axis y line*=none,
    axis x line*=bottom,
    yticklabels from table={\tqatransferdata}{Label}  
    ]
    \addlegendimage{empty legend}\addlegendentry{TriviaQA}
    
    \node [white, left] at (axis cs: 70.74,1) {+1.0};
    \node [white, left] at (axis cs: 67.6,2) {+1.0};
    \node [white, left] at (axis cs: 68.52,2) {+1.0};
    
    \addplot [style={transferbasecolor,fill=transferbasecolor,mark=none}] table [x=Baseline, meta=Label,y expr=\coordindex] {\tqatransferdata};    
    \addplot [style={transferlivecolor,fill=transferlivecolor,mark=none}] table [x=Live, meta=Label,y expr=\coordindex] {\tqatransferdata};
    \addplot [style={transfermemorycolor,fill=transfermemorycolor,mark=none}] table [x=Memory, meta=Label,y expr=\coordindex] {\tqatransferdata};
    \end{axis}
    \end{tikzpicture}
     \end{subfigure}
     \hfill
     \begin{subfigure}[t]{0.49\textwidth}
    \pgfplotstableread{ 
Label    Baseline Live Memory
FiD   49.03 0.0  0.0
L$\nicefrac{1}{3}$ 45.43    3.6  0.83
L$\nicefrac{1}{8}$ 40.17     4.98 1.39
    }\wqtransferdata     
    \begin{tikzpicture}

    \begin{axis}[
    xbar stacked,   
    ytick=data,     
    y dir=reverse,
    enlarge y limits=0.25,
    bar width=14pt,
    height = 5cm,
    width = \textwidth,
    legend pos={north west},
    axis y line*=none,
    axis x line*=bottom,
    yticklabels from table={\wqtransferdata}{Label}  
    ]
    \node [white, left] at (axis cs: 49.03,1) {+3.6};
    \node [white, left] at (axis cs: 50.05,1) {0.8};    
    \node [white, left] at (axis cs: 45.15,2) {+5.0};
    \node [white, left] at (axis cs: 46.54,2) {+1.4};
    
    \addlegendimage{empty legend}\addlegendentry{WebQuestions}
    \addplot [style={transferbasecolor,fill=transferbasecolor,mark=none}] table [x=Baseline, meta=Label,y expr=\coordindex] {\wqtransferdata};    
    \addplot [style={transferlivecolor,fill=transferlivecolor,mark=none}] table [x=Live, meta=Label,y expr=\coordindex] {\wqtransferdata};
    \addplot [style={transfermemorycolor,fill=transfermemorycolor,mark=none}] table [x=Memory, meta=Label,y expr=\coordindex] {\wqtransferdata};
    \end{axis}
    \end{tikzpicture}     
     \end{subfigure}
     \hfill
     
     \begin{tikzpicture}
        \begin{customlegend}[
            legend columns=3,
            legend style={
                align=center,
                column sep=4ex,
                font=\large,
            },
            legend entries={Baseline, Live, Memory}
        ]
            \addlegendimage{fill=transferbasecolor,color=transferbasecolor, line width=8pt}
            \addlegendimage{fill=transferlivecolor,color=transferlivecolor, line width=8pt}
            \addlegendimage{fill=transfermemorycolor,color=transfermemorycolor, line width=8pt}
        \end{customlegend}
    \end{tikzpicture}            
\caption{\textbf{Transferring memory and especially live encoder from a related dataset can partially close the gap with FiD, with increased gains for lower live proportion and smaller final dataset.} Exact match on TriviaQA and WebQuestions dev sets with and without transfer from Natural Questions for FiD and \modelname XL models with live proportion $\nicefrac{1}{3}$ and $\nicefrac{1}{8}$. Live keeps the memory encoder frozen during training on Natural Questions while Memory also trains the memory on Natural Questions (still frozen after transfer). The gains from transfer are much more pronounced for smaller live proportion and on WebQuestions, the smaller dataset.}
\label{fig:transfer}
\end{figure*}
\subsection{Main results}

Figure \ref{fig:perf_vs_liveprop} shows \modelname performance as a function of live proportion for varying model sizes. The first key observation is that a relatively small proportion of live layers is sufficient to achieve quality close to FiD. The second key observation is that as the model size increases, the required live proportion to recover FiD performance decreases. This pattern is further supported by results from Figure \ref{fig:gap_scale}, which explicitly measures how much of the gap between \memoryname and FiD is closed by \modelname and shows this gap increases with scale.

Figure \ref{fig:perf_vs_time} compares FLOPs as a function of performance for \modelname and FiD, demonstrating that \modelname achieves similar performance at lower FLOPs for fine-tuning and inference (assuming pre-computation is sufficiently amortized to be effectively free). Moreover, the advantage becomes more pronounced with larger model size, consistent with the findings from Figure \ref{fig:perf_vs_liveprop} and \ref{fig:gap_scale}. Figure \ref{fig:perf_vs_time_mem} shows that \modelname also has much stronger performance than \memoryname for any FLOP value. Finally, Table \ref{table:test_results} compares \modelname with published results in the literature.

\subsection{Transfer}
\label{section:transfer}

\begin{figure*}[t!]
     \centering
     \begin{subfigure}[t]{0.48\textwidth}
        \raggedright
        \begin{tikzpicture}[scale=1.0]
            \begin{axis}[
            scale only axis,
            width=0.9\columnwidth,
            height=0.43\textwidth,            
            ylabel={Exact Match},
            xlabel={\propaxis},
            mark=x,
            ymajorgrids=true,
            xmajorgrids=true,
            xminorticks=true,
            grid style=dashed,
            legend columns=1,
            legend cell align=left,
            legend pos={south east},
            ]
            \addlegendimage{empty legend}\addlegendentry{NaturalQ}
            \addplot[color=modelcolor,mark=\modelmark,mark size=1pt,line width=2] table {
                0 38.34
                0.041666 39.75
                0.125 42.78
                0.25 48.76
                0.333 50.46
                0.5 51.2
            };
            \addplot[color=finetunecolor,mark=\finetunemark,mark size=1pt,line width=2] table {
                0 43.47
                0.041666 43.61
                0.125 44.73
                0.25 49.79
                0.333 51.04
                0.5 51.7
            };
            \addplot[color=conditioncolor,mark=\conditionmark,mark size=1pt,line width=2] table {
                0.041666 43.09
                0.125 47.54
                0.25 49.95
                0.333 50.69
                0.5 50.89
            };
            \addplot[line width=1, dotted] table {
                0 51.79
                0.5 51.79
            };            
            \end{axis}
        \end{tikzpicture}  
     \end{subfigure}
     \hfill
     \begin{subfigure}[t]{0.48\textwidth}
        \centering
        \begin{tikzpicture}[scale=1.0]
            \begin{axis}[
            scale only axis,
            width=0.9\columnwidth,
            height=0.45\textwidth,
            xlabel={\propaxis},
            mark=x,
            ymajorgrids=true,
            xmajorgrids=true,
            xminorticks=true,
            grid style=dashed,
            legend columns=1,
            legend cell align=left,
            legend pos={south east},
            ]
            \addlegendimage{empty legend}\addlegendentry{TriviaQA}
            \addplot[color=modelcolor,mark=\modelmark,mark size=1pt,line width=2] table {
                0 55.05  
                0.041666 57.3
                0.125 61.94
                0.25 65.96
                0.333 67.16
                0.5 67.84
            };
            \addplot[color=finetunecolor,mark=\finetunemark,mark size=1pt,line width=2] table {
                0 60.78 
                0.041666 60.57
                0.25 66.76
                0.333 67.44
                0.5 67.84
            };
            \addplot[color=conditioncolor,mark=\conditionmark,mark size=1pt,line width=2] table {
                0 56.89 
                0.041666 62.37
                0.125 66.09
                0.25 67.47
                0.333 67.59
                0.5 67.84
            };
                        \addplot[line width=1, dotted] table {
                0 67.64
                0.5 67.64
            };         
            \end{axis}
        \end{tikzpicture}  
     \end{subfigure}     
     \hfill
     \begin{tikzpicture}
        \begin{customlegend}[
            legend columns=4,
            legend style={
                align=center,
                column sep=3ex,
                font=\large,
            },
            legend entries={\modelname, Fine-tune memory, Condition memory, FiD}
        ]
            \addlegendimage{mark=\modelmark,solid,color=modelcolor,line width=2}
            \addlegendimage{mark=\finetunemark,mark size=3pt,solid,color=finetunecolor,line width=2}
            \addlegendimage{mark=\conditionmark,mark size=3pt,solid,color=conditioncolor,line width=2}
            \addlegendimage{dotted,line width=1}
        \end{customlegend}
    \end{tikzpicture}            
     
    \caption{\textbf{Neither conditioning memory on input nor fine-tuning memory are sufficient to recover FiD performance. Both ingredients are important, although conditioning appears to contribute more.} Exact match on Natural Questions (NaturalQ) and TriviaQA dev sets as a function of proportion of live encoder layers for \modelname-Large and two relaxations: one in which the memory layers are fine-tuned, and another in which the memory layers are conditioned on the question.}
    \label{fig:perf_vs_prop_ablation}
\end{figure*}
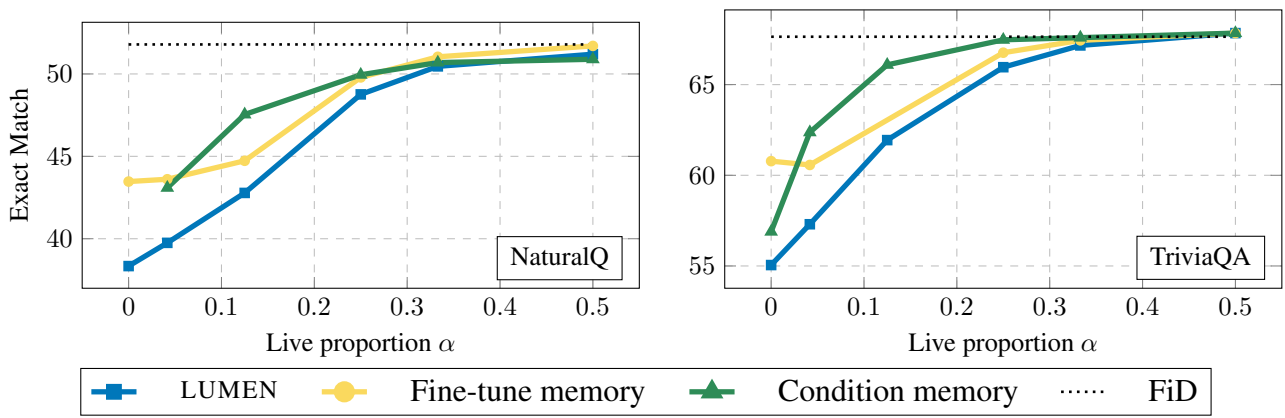
Since the memory encoder is not fine-tuned on each individual task, the live encoder must adapt the memory representations to the task in addition to conditioning on the input. Especially for smaller live encoders, this may be difficult to learn while fine-tuning on a single task. Here we evaluate whether \modelname can benefit from transferring from other knowledge-intensive tasks. 

In particular, we consider two transfer settings. In the \textit{Live} setting, we transfer the Live Encoder by training on Natural Questions with frozen memory before transferring to the target task. In the \textit{Memory} setting, the model is trained on Natural Questions with fine-tuned memory before transferring both the Live and Memory encoder to the target task. The \textit{Memory} setting follows the intuition that, although it is infeasible to use a different memory for every task, it may be possible to perform multi-task fine-tuning before encoding memory.

Figure \ref{fig:transfer} shows the results of transfer from Natural Questions to TriviaQA and WebQuestions. We note several interesting patterns. \textit{First}, gains from transfer are higher for smaller live proportion, with minimal gains for FiD and large gains for \modelname~$\nicefrac{1}{8}$. \textit{Second}, transferring memory is only helpful for small live proportion, where the Live Encoder does not contain sufficient layers to fully adapt the memory to the task. \textit{Third}, gains from transfer are significantly higher for WebQuestions, a task with a very small amount of data.

\subsection{Memory shape}
\label{section:mem_shape}

In our main experiments we initialize \modelname from public T5 checkpoints to avoid costly pre-training and partition the encoder into a memory encoder and live encoder. Can we achieve a better a trade-off by pre-training a model with a custom configuration? Fixing the output of the live encoder to have low model dimension allows us to \textit{scale the memory encoder without using more FLOPs}, as the cross-attention FLOPs are not affected by the size of the memory encoder. Table~\ref{table:memory_scale} shows the effect of adding a memory encoder consisting of 24 additional Base layers to an existing T5-Base configuration, yielding increasing performance without increasing compute. Taken to an extreme, these results suggest that combining a large language model with a moderately sized live encoder could yield strong results at modest cost.
\begin{table}[ht!]
\centering
\caption{\textbf{Adding memory to FiD leads to significant performance gains without additional fine-tuning or inference FLOPs.} Exact match performance on Natural Questions and TriviaQA for FiD-Base and \modelnameonethird with Base decoder and live encoder, and memory encoder with 24 Base layers.}
\vspace{0.35cm}
\begin{tabular}{l|cc}
    \textbf{Model} & \textbf{NQ} & \textbf{TQA}\\
    \toprule
    FiD Base  & 47.3 & 64.4 \\    
    \modelname Base 24-12 & \textbf{48.9} & \textbf{65.4} \\
    \bottomrule
\end{tabular}
\label{table:memory_scale}
\end{table}
\subsection{Ablations}

The two main differences between FiD, \modelname, and \memoryname are the extent to which retrieved passages are conditioned on the input and the extent to which passage encoders are fine-tuned on particular tasks. Our first ablation investigates how performance differences between \modelname and \memoryname on the one hand and FiD on the other hand result from conditioning on the input and fine-tuning. We construct two ablation settings as intermediate models between \modelname and FiD: fine-tuning the memory encoder, and conditioning the memory encoder on the question (but without fine-tuning it). Figure \ref{fig:perf_vs_prop_ablation} compares performance as a function of live proportion for these settings. Neither conditioning memory on the input nor fine-tuning the memory come close to recovering FiD performance by themselves: both are necessary. However, it seems that conditioning may be more helpful by itself than fine-tuning memory.

\begin{figure}[h!]
     \centering
    \hspace{-5pt}     
        \begin{tikzpicture}[scale=1.0]
            \begin{axis}[
            scale only axis,
            width=0.85\columnwidth,
            height=0.42\columnwidth,           
            ylabel={Exact Match},
            xlabel={\propaxis},
            mark=x,
            ymajorgrids=true,
            xmajorgrids=true,
            xminorticks=true,
            grid style=dashed,
            legend columns=1,
            legend cell align=left,
            legend pos={south east},
            ]
            \legend{\modelname, no q2mem, no mem2q}        
            \addplot[color=modelcolor,mark=\modelmark,mark size=1pt,line width=2] table {
                0 38.34
                0.041666 39.75
                0.125 42.78
                0.25 48.76
                0.333 50.46
                0.5 51.2
                0.75 51.52
            	1.0 51.79
            };
            \addplot[color=q2memcolor,mark=square,mark size=1pt,line width=2] table {
                0 38.11
                0.041666 39.73
                0.125 42.5
                0.25 49.31
                0.333 49.35
                0.5 50.03
                0.75 49.79
                1.0 49.14
            };
            \addplot[color=mem2qcolor,mark=star,mark size=1pt,line width=2] table {
                0 38.27
                0.041666 40.16
                0.125 41.36
                0.25 42.7
                0.333 43.15
                0.5 44.86
                0.75 45.55
                1.0 45.81
            };
            \end{axis}
        \end{tikzpicture}  
    \caption{\textbf{The primary gains from the live encoder in \modelname result from updating memory representations conditioned on the question.} Exact match on Natural Question dev set as a function of the proportion of live encoder layers for \modelname-Large and two modifications with restricted encoder self-attention. In the `no q2mem` setting question tokens cannot attend to passage tokens, and vice versa for `no mem2q`.}
    \label{fig:perf_vs_self_attention_masking}
\end{figure}
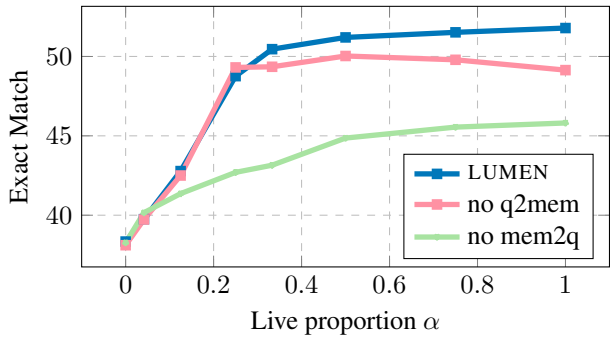

The \modelname live encoder jointly processes concatenated passage and input representations. The decoder therefore receives passages conditioned on the input as well as the input on the passage. In order to disentangle these conditioning effects, we experiment with ablations that disallows question tokens to attend to the passage (``no q2mem'') or passage tokens to question (``no mem2q''). Figure~\ref{fig:perf_vs_self_attention_masking} presents results that show that conditioning the passage on the input is critical, although the passage-conditioned question is still helpful.

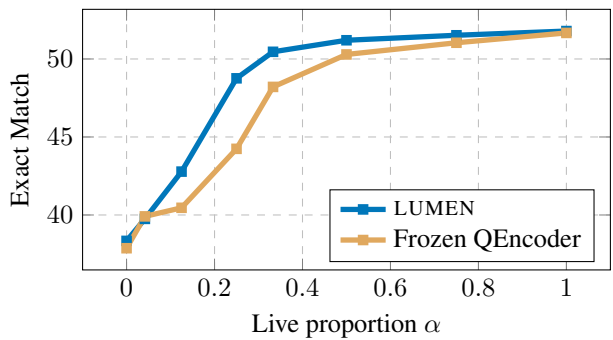
\begin{figure}[h!]
    \hspace{-5pt}
        \begin{tikzpicture}[scale=1.0]
            \begin{axis}[
            scale only axis,
            width=0.85\columnwidth,
            height=0.42\columnwidth,            
            ylabel={Exact Match},
            xlabel={\propaxis},
            mark=x,
            ymajorgrids=true,
            xmajorgrids=true,
            xminorticks=true,
            grid style=dashed,
            legend columns=1,
            legend cell align=left,
            legend pos={south east},
            ]
            \legend{\modelname, Frozen QEncoder}        
            \addplot[color=modelcolor,mark=\modelmark,mark size=1pt,line width=2] table {
                0 38.34
                0.041666 39.75
                0.125 42.78
                0.25 48.76
                0.333 50.46
                0.5 51.2
                0.75 51.52
            	1.0 51.79
            };
            \addplot[color=frozenqcolor,mark=square,mark size=1pt,line width=2] table {
                0 37.87
                0.041666 39.91
                0.125 40.46
                0.25 44.24
                0.333 48.21
                0.5 50.29
                0.75 51.04
                1.0 51.67
            };
            \end{axis}
        \end{tikzpicture}  
    \caption{\textbf{Fine-tuning the question encoder improves performance significantly.} Exact match on Natural Question dev set as a function of the proportion of live encoder layers for \modelname-Large and a modification for which the question encoder is frozen (so that the memory encoder and question encoder are shared).}
    \label{fig:perf_vs_prop_qenc}
\end{figure}

Finally, \modelname also uses a fine-tuned question encoder to generate a question representation that is optimized for the live encoder to condition the passage memories on. Figure \ref{fig:perf_vs_prop_qenc} compares performance between fine-tuning and freezing this question encoder, demonstrating the importance of adapting the question encoder to the task. 

\begin{table}[ht!]
\centering
\caption{Comparison of \modelname with published results on Natural Questions and TriviaQA test sets. We focus on comparing with FiD as other works enhance performance with improved retrieval (such as ATLAS), which is orthogonal and complementary to our contributions. \modelname is agnostic to retrieval method, and can be used with ATLAS retrieval.}
\vspace{0.35cm}
\begin{tabular}{l|cc}
    \textbf{Model} & \textbf{NQ} & \textbf{TQA}\\
    \toprule

    REALM \citep{realm} & 40.4 & - \\    
    RAG \citep{rag} & 44.5 & 56.8 \\
    RETRO \citep{retro} & 45.5 & - \\
    T5-XXL \citep{t5ssm} & 35.2 & 51.9 \\
    ATLAS \citep{atlas}  & 60.4 & 79.8 \\    
    \midrule
    FiD-L \citep{fid}  & 51.4 & 67.6 \\
    FiD-XXL (ours) & 57.3 & 73.0 \\
    \modelname-XXL  & 57.1 & 73.1 \\
    \bottomrule
\end{tabular}

\label{table:test_results}
\end{table}

\vspace{-0.4cm}
\section{Related Work}

\paragraph{Retrieval-augmented models}

There is a significant amount of research on retrieval-augmented language models. Some notable approaches include REALM~\citep{realm}, RAG~\citep{rag}, kNN-LM~\citep{knnlm}, RETRO~\citep{retro}, and Fusion-in-Decoder (FiD) \citep{fid}. FiD in particular has demonstrated state of the art performance across a range of tasks \citep{fid, atlas, generateratherretrieve}. This work focuses on improving the efficiency of FiD through a hybrid memory approach.

\paragraph{Efficient retrieval-augmented models}

Retrieval augmentation can be expensive for training and inference, and a large body of work investigates more efficient retrieval-augmented models. The computational cost of retrieval-augmented models can be partitioned into the cost from reading retrieved passages, decoding, and long-range attention. Recent work has shown that FiD spends the majority of inference time in the decoder ~\citep{fidlight} due to memory bandwidth constraints in cross-attention~\citep{fido}. However, with the appropriate modifications~\citep{fido} the constraint can be ameliorated, after which the majority of training and inference costs result from reading retrieved passages.

The computational burden from encoding retrieved passages can be reduced by reranking and making use of only the best retrievals~\citep{kgfid, r3rerank, readerguidererank}. Alternatively, the resources devoted to retrieval can be adapted to the difficulty of the input, retrieving fewer or no passages if the model is confident it already knows the answer~\citep{adaptiveretrieval, canext}. In order to efficiently model interaction between different retrieved passages it is common to employ sparse long-range attention~\citep{longt5, etc, readtwice}. Finally, there is a large body of work that attempts to improve the efficiency of Transformer models in general. Efficient fine-tuning methods~\citep{bitfit, lora} update only a fraction of parameters during fine-tuning, but unlike \modelname these methods still employ the full model for inference. Other general efficiency improvements include parallelization~\citep{palminference}, quantization~\citep{int8,glm}, and distillation~\citep{distill, distillsurvey}.

\paragraph{Memory models}

\modelname is most nearly related to the literature on \textit{memory}. Another method to reduce encoding cost of retrieval-augmented models is to pre-compute representations for the retrieval corpus and collect these representations into a memory, thereby amortizing the encoding cost over all the instances for which a sample is retrieved. In particular, \modelname is closely connected to \citet{fidmemory}, who propose a memory FiD model with pre-computed encoder representations. \modelname can be seen as a hybrid of \citet{fidmemory} and FiD that partially pre-computes encoder representations for efficiency, and finalizes the encoder representations on-the-fly conditioned on question and task to avoid the strong performance penalty from pre-computation. This partial pre-computation is a form of late interaction, which has previously been used to improve the selection of retrieved passages~\citep{colbert, colbertv2}. We instead employ late interaction in the reader model. \citet{lait} also employs late interaction in the reader model, but in the context of NLI as opposed to retrieval-augmented generation.

\modelname uses memory in a straightforward manner, simply pre-computing token representations from a pre-trained model and retrieving passages with a standard dense passage retriever. Other memory models can be more involved, incorporating end-to-end retrieval within the model ~\citep{tome, memorizing}, storing higher-level latent representations~\citep{tome, qama, emat}, and specific pre-training for memory~\citep{tome, trime}. The main idea behind \modelname to update retrieved memory representations conditioning on the input is complementary to and can be combined with these more complex memory models.

\section{Conclusion}

Retrieval-augmented language models such as Fusion-in-Decoder are powerful but expensive. Pre-computing encoder representations into dense memory, a popular method for reducing computation costs of retrieval-augmented models, leads to a sharp decrease in performance. We propose \modelname, a hybrid between Fusion-in-Decoder and dense memory. Passage representations are partially pre-encoded into a dense memory, and then reprocessed on the fly by a fine-tuned encoder that conditions on the question. We show that \modelname achieves stronger performance for the same FLOPs, and that this advantage increases with scale.


\section*{Acknowlegements}
We thank DeLesley Hutchins, Santiago Ontanon, Pat Verga, Markus Rabe, Yuhai Wu, Emma Strubell, and others at Google Research for insightful advice and discussion. Michiel de Jong is partially supported by NSF Awards IIS-1513966/ 1632803/1833137, CCF-1139148, DARPA Awards\#: FA8750-18-2-0117, FA8750-19-1-0504,  DARPA-D3M - Award UCB-00009528, Google Research Awards, gifts from Facebook and Netflix, and ARO\# W911NF-12-1-0241 and W911NF-15-1-0484.

\bibliography{custom}

\begin{thebibliography}{46}
\providecommand{\natexlab}[1]{#1}
\providecommand{\url}[1]{\texttt{#1}}
\expandafter\ifx\csname urlstyle\endcsname\relax
  \providecommand{\doi}[1]{doi: #1}\else
  \providecommand{\doi}{doi: \begingroup \urlstyle{rm}\Url}\fi

\bibitem[Ainslie et~al.(2020)Ainslie, Onta{\~{n}}{\'{o}}n, Alberti, Cvicek,
  Fisher, Pham, Ravula, Sanghai, Wang, and Yang]{etc}
Ainslie, J., Onta{\~{n}}{\'{o}}n, S., Alberti, C., Cvicek, V., Fisher, Z.,
  Pham, P., Ravula, A., Sanghai, S., Wang, Q., and Yang, L.
\newblock {ETC:} encoding long and structured inputs in transformers.
\newblock In Webber, B., Cohn, T., He, Y., and Liu, Y. (eds.),
  \emph{Proceedings of the 2020 Conference on Empirical Methods in Natural
  Language Processing, {EMNLP} 2020, Online, November 16-20, 2020}, pp.\
  268--284. Association for Computational Linguistics, 2020.
\newblock \doi{10.18653/v1/2020.emnlp-main.19}.
\newblock URL \url{https://doi.org/10.18653/v1/2020.emnlp-main.19}.

\bibitem[Ainslie et~al.(2023)Ainslie, Lee-Thorp, de~Jong, Zemlyanskiy, Lebrón,
  and Sanghai]{gqa}
Ainslie, J., Lee-Thorp, J., de~Jong, M., Zemlyanskiy, Y., Lebrón, F., and
  Sanghai, S.
\newblock Gqa: Training generalized multi-query transformer models from
  multi-head checkpoints, 2023.

\bibitem[Berant et~al.(2013)Berant, Chou, Frostig, and Liang]{webquestions}
Berant, J., Chou, A., Frostig, R., and Liang, P.
\newblock Semantic parsing on freebase from question-answer pairs.
\newblock In \emph{Proceedings of the 2013 Conference on Empirical Methods in
  Natural Language Processing, {EMNLP} 2013, 18-21 October 2013, Grand Hyatt
  Seattle, Seattle, Washington, USA, {A} meeting of SIGDAT, a Special Interest
  Group of the {ACL}}, pp.\  1533--1544. {ACL}, 2013.
\newblock URL \url{https://aclanthology.org/D13-1160/}.

\bibitem[Borgeaud et~al.(2022)Borgeaud, Mensch, Hoffmann, Cai, Rutherford,
  Millican, van~den Driessche, Lespiau, Damoc, Clark, de~Las~Casas, Guy,
  Menick, Ring, Hennigan, Huang, Maggiore, Jones, Cassirer, Brock, Paganini,
  Irving, Vinyals, Osindero, Simonyan, Rae, Elsen, and Sifre]{retro}
Borgeaud, S., Mensch, A., Hoffmann, J., Cai, T., Rutherford, E., Millican, K.,
  van~den Driessche, G., Lespiau, J., Damoc, B., Clark, A., de~Las~Casas, D.,
  Guy, A., Menick, J., Ring, R., Hennigan, T., Huang, S., Maggiore, L., Jones,
  C., Cassirer, A., Brock, A., Paganini, M., Irving, G., Vinyals, O., Osindero,
  S., Simonyan, K., Rae, J.~W., Elsen, E., and Sifre, L.
\newblock Improving language models by retrieving from trillions of tokens.
\newblock In Chaudhuri, K., Jegelka, S., Song, L., Szepesv{\'{a}}ri, C., Niu,
  G., and Sabato, S. (eds.), \emph{International Conference on Machine
  Learning, {ICML} 2022, 17-23 July 2022, Baltimore, Maryland, {USA}}, volume
  162 of \emph{Proceedings of Machine Learning Research}, pp.\  2206--2240.
  {PMLR}, 2022.
\newblock URL \url{https://proceedings.mlr.press/v162/borgeaud22a.html}.

\bibitem[Bradbury et~al.(2018)Bradbury, Frostig, Hawkins, Johnson, Leary,
  Maclaurin, Necula, Paszke, Vander{P}las, Wanderman-{M}ilne, and Zhang]{jax}
Bradbury, J., Frostig, R., Hawkins, P., Johnson, M.~J., Leary, C., Maclaurin,
  D., Necula, G., Paszke, A., Vander{P}las, J., Wanderman-{M}ilne, S., and
  Zhang, Q.
\newblock {JAX}: composable transformations of {P}ython+{N}um{P}y programs,
  2018.
\newblock URL \url{http://github.com/google/jax}.

\bibitem[Chen et~al.(2022)Chen, Verga, de~Jong, Wieting, and Cohen]{qama}
Chen, W., Verga, P., de~Jong, M., Wieting, J., and Cohen, W.~W.
\newblock Augmenting pre-trained language models with qa-memory for open-domain
  question answering.
\newblock \emph{CoRR}, abs/2204.04581, 2022.
\newblock \doi{10.48550/arXiv.2204.04581}.
\newblock URL \url{https://doi.org/10.48550/arXiv.2204.04581}.

\bibitem[de~Jong et~al.(2022{\natexlab{a}})de~Jong, Zemlyanskiy, Ainslie,
  FitzGerald, Sanghai, Sha, and Cohen]{fido}
de~Jong, M., Zemlyanskiy, Y., Ainslie, J., FitzGerald, N., Sanghai, S., Sha,
  F., and Cohen, W.
\newblock Fido: Fusion-in-decoder optimized for stronger performance and faster
  inference.
\newblock \emph{arXiv preprint arXiv:2212.08153}, 2022{\natexlab{a}}.

\bibitem[de~Jong et~al.(2022{\natexlab{b}})de~Jong, Zemlyanskiy, FitzGerald,
  Sha, and Cohen]{tome}
de~Jong, M., Zemlyanskiy, Y., FitzGerald, N., Sha, F., and Cohen, W.~W.
\newblock Mention memory: incorporating textual knowledge into transformers
  through entity mention attention.
\newblock In \emph{The Tenth International Conference on Learning
  Representations, {ICLR} 2022, Virtual Event, April 25-29, 2022}.
  OpenReview.net, 2022{\natexlab{b}}.
\newblock URL \url{https://openreview.net/forum?id=OY1A8ejQgEX}.

\bibitem[Dettmers et~al.(2022)Dettmers, Lewis, Belkada, and Zettlemoyer]{int8}
Dettmers, T., Lewis, M., Belkada, Y., and Zettlemoyer, L.
\newblock Llm.int8(): 8-bit matrix multiplication for transformers at scale.
\newblock \emph{CoRR}, abs/2208.07339, 2022.
\newblock \doi{10.48550/arXiv.2208.07339}.
\newblock URL \url{https://doi.org/10.48550/arXiv.2208.07339}.

\bibitem[Google(2022)]{t5check}
Google.
\newblock Pre-trained t5 models.
\newblock
  \url{https://github.com/google-research/t5x/blob/main/docs/models.md}, 2022.
\newblock Accessed: 2022-12-20.

\bibitem[Gou et~al.(2021)Gou, Yu, Maybank, and Tao]{distillsurvey}
Gou, J., Yu, B., Maybank, S.~J., and Tao, D.
\newblock Knowledge distillation: {A} survey.
\newblock \emph{Int. J. Comput. Vis.}, 129\penalty0 (6):\penalty0 1789--1819,
  2021.
\newblock \doi{10.1007/s11263-021-01453-z}.
\newblock URL \url{https://doi.org/10.1007/s11263-021-01453-z}.

\bibitem[Guo et~al.(2022)Guo, Ainslie, Uthus, Onta{\~{n}}{\'{o}}n, Ni, Sung,
  and Yang]{longt5}
Guo, M., Ainslie, J., Uthus, D.~C., Onta{\~{n}}{\'{o}}n, S., Ni, J., Sung, Y.,
  and Yang, Y.
\newblock Longt5: Efficient text-to-text transformer for long sequences.
\newblock In Carpuat, M., de~Marneffe, M., and Ru{\'{\i}}z, I. V.~M. (eds.),
  \emph{Findings of the Association for Computational Linguistics: {NAACL}
  2022, Seattle, WA, United States, July 10-15, 2022}, pp.\  724--736.
  Association for Computational Linguistics, 2022.
\newblock \doi{10.18653/v1/2022.findings-naacl.55}.
\newblock URL \url{https://doi.org/10.18653/v1/2022.findings-naacl.55}.

\bibitem[Guu et~al.(2020)Guu, Lee, Tung, Pasupat, and Chang]{realm}
Guu, K., Lee, K., Tung, Z., Pasupat, P., and Chang, M.
\newblock {REALM:} retrieval-augmented language model pre-training.
\newblock \emph{CoRR}, abs/2002.08909, 2020.
\newblock URL \url{https://arxiv.org/abs/2002.08909}.

\bibitem[Heek et~al.(2020)Heek, Levskaya, Oliver, Ritter, Rondepierre, Steiner,
  and van {Z}ee]{flax}
Heek, J., Levskaya, A., Oliver, A., Ritter, M., Rondepierre, B., Steiner, A.,
  and van {Z}ee, M.
\newblock {F}lax: A neural network library and ecosystem for {JAX}, 2020.
\newblock URL \url{http://github.com/google/flax}.

\bibitem[Hinton et~al.(2015)Hinton, Vinyals, and Dean]{distill}
Hinton, G.~E., Vinyals, O., and Dean, J.
\newblock Distilling the knowledge in a neural network.
\newblock \emph{CoRR}, abs/1503.02531, 2015.
\newblock URL \url{http://arxiv.org/abs/1503.02531}.

\bibitem[Hofst{\"{a}}tter et~al.(2022)Hofst{\"{a}}tter, Chen, Raman, and
  Zamani]{fidlight}
Hofst{\"{a}}tter, S., Chen, J., Raman, K., and Zamani, H.
\newblock Fid-light: Efficient and effective retrieval-augmented text
  generation.
\newblock \emph{CoRR}, abs/2209.14290, 2022.
\newblock \doi{10.48550/arXiv.2209.14290}.
\newblock URL \url{https://doi.org/10.48550/arXiv.2209.14290}.

\bibitem[Hu et~al.(2022)Hu, Shen, Wallis, Allen{-}Zhu, Li, Wang, Wang, and
  Chen]{lora}
Hu, E.~J., Shen, Y., Wallis, P., Allen{-}Zhu, Z., Li, Y., Wang, S., Wang, L.,
  and Chen, W.
\newblock Lora: Low-rank adaptation of large language models.
\newblock In \emph{The Tenth International Conference on Learning
  Representations, {ICLR} 2022, Virtual Event, April 25-29, 2022}.
  OpenReview.net, 2022.
\newblock URL \url{https://openreview.net/forum?id=nZeVKeeFYf9}.

\bibitem[Izacard \& Grave(2021)Izacard and Grave]{fid}
Izacard, G. and Grave, E.
\newblock Leveraging passage retrieval with generative models for open domain
  question answering.
\newblock In Merlo, P., Tiedemann, J., and Tsarfaty, R. (eds.),
  \emph{Proceedings of the 16th Conference of the European Chapter of the
  Association for Computational Linguistics: Main Volume, {EACL} 2021, Online,
  April 19 - 23, 2021}, pp.\  874--880. Association for Computational
  Linguistics, 2021.
\newblock \doi{10.18653/v1/2021.eacl-main.74}.
\newblock URL \url{https://doi.org/10.18653/v1/2021.eacl-main.74}.

\bibitem[Izacard et~al.(2022)Izacard, Lewis, Lomeli, Hosseini, Petroni, Schick,
  Dwivedi{-}Yu, Joulin, Riedel, and Grave]{atlas}
Izacard, G., Lewis, P., Lomeli, M., Hosseini, L., Petroni, F., Schick, T.,
  Dwivedi{-}Yu, J., Joulin, A., Riedel, S., and Grave, E.
\newblock Few-shot learning with retrieval augmented language models.
\newblock \emph{CoRR}, abs/2208.03299, 2022.
\newblock \doi{10.48550/arXiv.2208.03299}.
\newblock URL \url{https://doi.org/10.48550/arXiv.2208.03299}.

\bibitem[Joshi et~al.(2017)Joshi, Choi, Weld, and Zettlemoyer]{triviaqa}
Joshi, M., Choi, E., Weld, D.~S., and Zettlemoyer, L.
\newblock Triviaqa: {A} large scale distantly supervised challenge dataset for
  reading comprehension.
\newblock In Barzilay, R. and Kan, M. (eds.), \emph{Proceedings of the 55th
  Annual Meeting of the Association for Computational Linguistics, {ACL} 2017,
  Vancouver, Canada, July 30 - August 4, Volume 1: Long Papers}, pp.\
  1601--1611. Association for Computational Linguistics, 2017.
\newblock \doi{10.18653/v1/P17-1147}.
\newblock URL \url{https://doi.org/10.18653/v1/P17-1147}.

\bibitem[Karpukhin et~al.(2020)Karpukhin, Oguz, Min, Lewis, Wu, Edunov, Chen,
  and Yih]{dpr}
Karpukhin, V., Oguz, B., Min, S., Lewis, P. S.~H., Wu, L., Edunov, S., Chen,
  D., and Yih, W.
\newblock Dense passage retrieval for open-domain question answering.
\newblock In Webber, B., Cohn, T., He, Y., and Liu, Y. (eds.),
  \emph{Proceedings of the 2020 Conference on Empirical Methods in Natural
  Language Processing, {EMNLP} 2020, Online, November 16-20, 2020}, pp.\
  6769--6781. Association for Computational Linguistics, 2020.
\newblock \doi{10.18653/v1/2020.emnlp-main.550}.
\newblock URL \url{https://doi.org/10.18653/v1/2020.emnlp-main.550}.

\bibitem[Khandelwal et~al.(2020)Khandelwal, Levy, Jurafsky, Zettlemoyer, and
  Lewis]{knnlm}
Khandelwal, U., Levy, O., Jurafsky, D., Zettlemoyer, L., and Lewis, M.
\newblock Generalization through memorization: Nearest neighbor language
  models.
\newblock In \emph{8th International Conference on Learning Representations,
  {ICLR} 2020, Addis Ababa, Ethiopia, April 26-30, 2020}. OpenReview.net, 2020.
\newblock URL \url{https://openreview.net/forum?id=HklBjCEKvH}.

\bibitem[Khattab \& Zaharia(2020)Khattab and Zaharia]{colbert}
Khattab, O. and Zaharia, M.
\newblock Colbert: Efficient and effective passage search via contextualized
  late interaction over {BERT}.
\newblock In Huang, J.~X., Chang, Y., Cheng, X., Kamps, J., Murdock, V., Wen,
  J., and Liu, Y. (eds.), \emph{Proceedings of the 43rd International {ACM}
  {SIGIR} conference on research and development in Information Retrieval,
  {SIGIR} 2020, Virtual Event, China, July 25-30, 2020}, pp.\  39--48. {ACM},
  2020.
\newblock \doi{10.1145/3397271.3401075}.
\newblock URL \url{https://doi.org/10.1145/3397271.3401075}.

\bibitem[Kratzwald \& Feuerriegel(2018)Kratzwald and
  Feuerriegel]{adaptiveretrieval}
Kratzwald, B. and Feuerriegel, S.
\newblock Adaptive document retrieval for deep question answering.
\newblock In Riloff, E., Chiang, D., Hockenmaier, J., and Tsujii, J. (eds.),
  \emph{Proceedings of the 2018 Conference on Empirical Methods in Natural
  Language Processing, Brussels, Belgium, October 31 - November 4, 2018}, pp.\
  576--581. Association for Computational Linguistics, 2018.
\newblock \doi{10.18653/v1/d18-1055}.
\newblock URL \url{https://doi.org/10.18653/v1/d18-1055}.

\bibitem[Kwiatkowski et~al.(2019)Kwiatkowski, Palomaki, Redfield, Collins,
  Parikh, Alberti, Epstein, Polosukhin, Devlin, Lee, Toutanova, Jones, Kelcey,
  Chang, Dai, Uszkoreit, Le, and Petrov]{nq}
Kwiatkowski, T., Palomaki, J., Redfield, O., Collins, M., Parikh, A.~P.,
  Alberti, C., Epstein, D., Polosukhin, I., Devlin, J., Lee, K., Toutanova, K.,
  Jones, L., Kelcey, M., Chang, M., Dai, A.~M., Uszkoreit, J., Le, Q., and
  Petrov, S.
\newblock Natural questions: a benchmark for question answering research.
\newblock \emph{Trans. Assoc. Comput. Linguistics}, 7:\penalty0 452--466, 2019.
\newblock \doi{10.1162/tacl\_a\_00276}.
\newblock URL \url{https://doi.org/10.1162/tacl\_a\_00276}.

\bibitem[Lewis et~al.(2020)Lewis, Perez, Piktus, Petroni, Karpukhin, Goyal,
  K{\"{u}}ttler, Lewis, Yih, Rockt{\"{a}}schel, Riedel, and Kiela]{rag}
Lewis, P. S.~H., Perez, E., Piktus, A., Petroni, F., Karpukhin, V., Goyal, N.,
  K{\"{u}}ttler, H., Lewis, M., Yih, W., Rockt{\"{a}}schel, T., Riedel, S., and
  Kiela, D.
\newblock Retrieval-augmented generation for knowledge-intensive {NLP} tasks.
\newblock In Larochelle, H., Ranzato, M., Hadsell, R., Balcan, M., and Lin, H.
  (eds.), \emph{Advances in Neural Information Processing Systems 33: Annual
  Conference on Neural Information Processing Systems 2020, NeurIPS 2020,
  December 6-12, 2020, virtual}, 2020.
\newblock URL
  \url{https://proceedings.neurips.cc/paper/2020/hash/6b493230205f780e1bc26945df7481e5-Abstract.html}.

\bibitem[Li et~al.(2022)Li, Guo, and Kumar]{fidmemory}
Li, Z., Guo, R., and Kumar, S.
\newblock Decoupled context processing for context augmented language modeling.
\newblock \emph{CoRR}, abs/2210.05758, 2022.
\newblock \doi{10.48550/arXiv.2210.05758}.
\newblock URL \url{https://doi.org/10.48550/arXiv.2210.05758}.

\bibitem[Mao et~al.(2021)Mao, He, Liu, Shen, Gao, Han, and
  Chen]{readerguidererank}
Mao, Y., He, P., Liu, X., Shen, Y., Gao, J., Han, J., and Chen, W.
\newblock Reader-guided passage reranking for open-domain question answering.
\newblock In Zong, C., Xia, F., Li, W., and Navigli, R. (eds.), \emph{Findings
  of the Association for Computational Linguistics: {ACL/IJCNLP} 2021, Online
  Event, August 1-6, 2021}, volume {ACL/IJCNLP} 2021 of \emph{Findings of
  {ACL}}, pp.\  344--350. Association for Computational Linguistics, 2021.
\newblock \doi{10.18653/v1/2021.findings-acl.29}.
\newblock URL \url{https://doi.org/10.18653/v1/2021.findings-acl.29}.

\bibitem[Milbauer et~al.(2023)Milbauer, Louis, Hosseini, Fabrikant, Metzler,
  and Schuster]{lait}
Milbauer, J.~L., Louis, A., Hosseini, J., Fabrikant, A., Metzler, D., and
  Schuster, T.
\newblock Lait: Efficient multi-segment encoding in transformers with
  layer-adjustable interaction.
\newblock In \emph{Proceedings of the Association for Computational
  Linguistics: ACL 2023}, 2023.

\bibitem[Pope et~al.(2022)Pope, Douglas, Chowdhery, Devlin, Bradbury, Levskaya,
  Heek, Xiao, Agrawal, and Dean]{palminference}
Pope, R., Douglas, S., Chowdhery, A., Devlin, J., Bradbury, J., Levskaya, A.,
  Heek, J., Xiao, K., Agrawal, S., and Dean, J.
\newblock Efficiently scaling transformer inference.
\newblock \emph{CoRR}, abs/2211.05102, 2022.
\newblock \doi{10.48550/arXiv.2211.05102}.
\newblock URL \url{https://doi.org/10.48550/arXiv.2211.05102}.

\bibitem[Raffel et~al.(2020)Raffel, Shazeer, Roberts, Lee, Narang, Matena,
  Zhou, Li, and Liu]{t5}
Raffel, C., Shazeer, N., Roberts, A., Lee, K., Narang, S., Matena, M., Zhou,
  Y., Li, W., and Liu, P.~J.
\newblock Exploring the limits of transfer learning with a unified text-to-text
  transformer.
\newblock \emph{J. Mach. Learn. Res.}, 21:\penalty0 140:1--140:67, 2020.
\newblock URL \url{http://jmlr.org/papers/v21/20-074.html}.

\bibitem[Roberts et~al.(2020)Roberts, Raffel, and Shazeer]{t5ssm}
Roberts, A., Raffel, C., and Shazeer, N.
\newblock How much knowledge can you pack into the parameters of a language
  model?
\newblock In Webber, B., Cohn, T., He, Y., and Liu, Y. (eds.),
  \emph{Proceedings of the 2020 Conference on Empirical Methods in Natural
  Language Processing, {EMNLP} 2020, Online, November 16-20, 2020}, pp.\
  5418--5426. Association for Computational Linguistics, 2020.
\newblock \doi{10.18653/v1/2020.emnlp-main.437}.
\newblock URL \url{https://doi.org/10.18653/v1/2020.emnlp-main.437}.

\bibitem[Roberts et~al.(2022)Roberts, Chung, Levskaya, Mishra, Bradbury, Andor,
  Narang, Lester, Gaffney, Mohiuddin, Hawthorne, Lewkowycz, Salcianu, van Zee,
  Austin, Goodman, Soares, Hu, Tsvyashchenko, Chowdhery, Bastings, Bulian,
  Garcia, Ni, Chen, Kenealy, Clark, Lee, Garrette, Lee-Thorp, Raffel, Shazeer,
  Ritter, Bosma, Passos, Maitin-Shepard, Fiedel, Omernick, Saeta, Sepassi,
  Spiridonov, Newlan, and Gesmundo]{t5x}
Roberts, A., Chung, H.~W., Levskaya, A., Mishra, G., Bradbury, J., Andor, D.,
  Narang, S., Lester, B., Gaffney, C., Mohiuddin, A., Hawthorne, C., Lewkowycz,
  A., Salcianu, A., van Zee, M., Austin, J., Goodman, S., Soares, L.~B., Hu,
  H., Tsvyashchenko, S., Chowdhery, A., Bastings, J., Bulian, J., Garcia, X.,
  Ni, J., Chen, A., Kenealy, K., Clark, J.~H., Lee, S., Garrette, D.,
  Lee-Thorp, J., Raffel, C., Shazeer, N., Ritter, M., Bosma, M., Passos, A.,
  Maitin-Shepard, J., Fiedel, N., Omernick, M., Saeta, B., Sepassi, R.,
  Spiridonov, A., Newlan, J., and Gesmundo, A.
\newblock Scaling up models and data with $\texttt{t5x}$ and $\texttt{seqio}$.
\newblock \emph{arXiv preprint arXiv:2203.17189}, 2022.
\newblock URL \url{https://arxiv.org/abs/2203.17189}.

\bibitem[Santhanam et~al.(2022)Santhanam, Khattab, Saad{-}Falcon, Potts, and
  Zaharia]{colbertv2}
Santhanam, K., Khattab, O., Saad{-}Falcon, J., Potts, C., and Zaharia, M.
\newblock Colbertv2: Effective and efficient retrieval via lightweight late
  interaction.
\newblock In Carpuat, M., de~Marneffe, M., and Ru{\'{\i}}z, I. V.~M. (eds.),
  \emph{Proceedings of the 2022 Conference of the North American Chapter of the
  Association for Computational Linguistics: Human Language Technologies,
  {NAACL} 2022, Seattle, WA, United States, July 10-15, 2022}, pp.\
  3715--3734. Association for Computational Linguistics, 2022.
\newblock \doi{10.18653/v1/2022.naacl-main.272}.
\newblock URL \url{https://doi.org/10.18653/v1/2022.naacl-main.272}.

\bibitem[Shazeer(2019)]{multiquery}
Shazeer, N.
\newblock Fast transformer decoding: One write-head is all you need.
\newblock \emph{CoRR}, abs/1911.02150, 2019.
\newblock URL \url{http://arxiv.org/abs/1911.02150}.

\bibitem[Shazeer \& Stern(2018)Shazeer and Stern]{adafactor}
Shazeer, N. and Stern, M.
\newblock Adafactor: Adaptive learning rates with sublinear memory cost.
\newblock In Dy, J.~G. and Krause, A. (eds.), \emph{Proceedings of the 35th
  International Conference on Machine Learning, {ICML} 2018,
  Stockholmsm{\"{a}}ssan, Stockholm, Sweden, July 10-15, 2018}, volume~80 of
  \emph{Proceedings of Machine Learning Research}, pp.\  4603--4611. {PMLR},
  2018.
\newblock URL \url{http://proceedings.mlr.press/v80/shazeer18a.html}.

\bibitem[Varshney et~al.(2022)Varshney, Luo, and Baral]{canext}
Varshney, N., Luo, M., and Baral, C.
\newblock Can open-domain {QA} reader utilize external knowledge efficiently
  like humans?
\newblock \emph{CoRR}, abs/2211.12707, 2022.
\newblock \doi{10.48550/arXiv.2211.12707}.
\newblock URL \url{https://doi.org/10.48550/arXiv.2211.12707}.

\bibitem[Wang et~al.(2018)Wang, Yu, Guo, Wang, Klinger, Zhang, Chang, Tesauro,
  Zhou, and Jiang]{r3rerank}
Wang, S., Yu, M., Guo, X., Wang, Z., Klinger, T., Zhang, W., Chang, S.,
  Tesauro, G., Zhou, B., and Jiang, J.
\newblock R\({}^{\mbox{3}}\): Reinforced ranker-reader for open-domain question
  answering.
\newblock In McIlraith, S.~A. and Weinberger, K.~Q. (eds.), \emph{Proceedings
  of the Thirty-Second {AAAI} Conference on Artificial Intelligence, (AAAI-18),
  the 30th innovative Applications of Artificial Intelligence (IAAI-18), and
  the 8th {AAAI} Symposium on Educational Advances in Artificial Intelligence
  (EAAI-18), New Orleans, Louisiana, USA, February 2-7, 2018}, pp.\
  5981--5988. {AAAI} Press, 2018.
\newblock URL
  \url{https://www.aaai.org/ocs/index.php/AAAI/AAAI18/paper/view/16712}.

\bibitem[Wu et~al.(2022{\natexlab{a}})Wu, Rabe, Hutchins, and
  Szegedy]{memorizing}
Wu, Y., Rabe, M.~N., Hutchins, D., and Szegedy, C.
\newblock Memorizing transformers.
\newblock In \emph{The Tenth International Conference on Learning
  Representations, {ICLR} 2022, Virtual Event, April 25-29, 2022}.
  OpenReview.net, 2022{\natexlab{a}}.
\newblock URL \url{https://openreview.net/forum?id=TrjbxzRcnf-}.

\bibitem[Wu et~al.(2022{\natexlab{b}})Wu, Zhao, Hu, Minervini, Stenetorp, and
  Riedel]{emat}
Wu, Y., Zhao, Y., Hu, B., Minervini, P., Stenetorp, P., and Riedel, S.
\newblock An efficient memory-augmented transformer for knowledge-intensive
  {NLP} tasks.
\newblock \emph{CoRR}, abs/2210.16773, 2022{\natexlab{b}}.
\newblock \doi{10.48550/arXiv.2210.16773}.
\newblock URL \url{https://doi.org/10.48550/arXiv.2210.16773}.

\bibitem[Yu et~al.(2022{\natexlab{a}})Yu, Zhu, Fang, Yu, Wang, Xu, Ren, Yang,
  and Zeng]{kgfid}
Yu, D., Zhu, C., Fang, Y., Yu, W., Wang, S., Xu, Y., Ren, X., Yang, Y., and
  Zeng, M.
\newblock Kg-fid: Infusing knowledge graph in fusion-in-decoder for open-domain
  question answering.
\newblock In Muresan, S., Nakov, P., and Villavicencio, A. (eds.),
  \emph{Proceedings of the 60th Annual Meeting of the Association for
  Computational Linguistics (Volume 1: Long Papers), {ACL} 2022, Dublin,
  Ireland, May 22-27, 2022}, pp.\  4961--4974. Association for Computational
  Linguistics, 2022{\natexlab{a}}.
\newblock \doi{10.18653/v1/2022.acl-long.340}.
\newblock URL \url{https://doi.org/10.18653/v1/2022.acl-long.340}.

\bibitem[Yu et~al.(2022{\natexlab{b}})Yu, Iter, Wang, Xu, Ju, Sanyal, Zhu,
  Zeng, and Jiang]{generateratherretrieve}
Yu, W., Iter, D., Wang, S., Xu, Y., Ju, M., Sanyal, S., Zhu, C., Zeng, M., and
  Jiang, M.
\newblock Generate rather than retrieve: Large language models are strong
  context generators.
\newblock \emph{CoRR}, abs/2209.10063, 2022{\natexlab{b}}.
\newblock \doi{10.48550/arXiv.2209.10063}.
\newblock URL \url{https://doi.org/10.48550/arXiv.2209.10063}.

\bibitem[Zaken et~al.(2022)Zaken, Goldberg, and Ravfogel]{bitfit}
Zaken, E.~B., Goldberg, Y., and Ravfogel, S.
\newblock Bitfit: Simple parameter-efficient fine-tuning for transformer-based
  masked language-models.
\newblock In Muresan, S., Nakov, P., and Villavicencio, A. (eds.),
  \emph{Proceedings of the 60th Annual Meeting of the Association for
  Computational Linguistics (Volume 2: Short Papers), {ACL} 2022, Dublin,
  Ireland, May 22-27, 2022}, pp.\  1--9. Association for Computational
  Linguistics, 2022.
\newblock \doi{10.18653/v1/2022.acl-short.1}.
\newblock URL \url{https://doi.org/10.18653/v1/2022.acl-short.1}.

\bibitem[Zemlyanskiy et~al.(2021)Zemlyanskiy, Ainslie, de~Jong, Pham, Eckstein,
  and Sha]{readtwice}
Zemlyanskiy, Y., Ainslie, J., de~Jong, M., Pham, P., Eckstein, I., and Sha, F.
\newblock Readtwice: Reading very large documents with memories.
\newblock In Toutanova, K., Rumshisky, A., Zettlemoyer, L.,
  Hakkani{-}T{\"{u}}r, D., Beltagy, I., Bethard, S., Cotterell, R.,
  Chakraborty, T., and Zhou, Y. (eds.), \emph{Proceedings of the 2021
  Conference of the North American Chapter of the Association for Computational
  Linguistics: Human Language Technologies, {NAACL-HLT} 2021, Online, June
  6-11, 2021}, pp.\  5189--5195. Association for Computational Linguistics,
  2021.
\newblock \doi{10.18653/v1/2021.naacl-main.408}.
\newblock URL \url{https://doi.org/10.18653/v1/2021.naacl-main.408}.

\bibitem[Zeng et~al.(2022)Zeng, Liu, Du, Wang, Lai, Ding, Yang, Xu, Zheng, Xia,
  Tam, Ma, Xue, Zhai, Chen, Zhang, Dong, and Tang]{glm}
Zeng, A., Liu, X., Du, Z., Wang, Z., Lai, H., Ding, M., Yang, Z., Xu, Y.,
  Zheng, W., Xia, X., Tam, W.~L., Ma, Z., Xue, Y., Zhai, J., Chen, W., Zhang,
  P., Dong, Y., and Tang, J.
\newblock {GLM-130B:} an open bilingual pre-trained model.
\newblock \emph{CoRR}, abs/2210.02414, 2022.
\newblock \doi{10.48550/arXiv.2210.02414}.
\newblock URL \url{https://doi.org/10.48550/arXiv.2210.02414}.

\bibitem[Zhong et~al.(2022)Zhong, Lei, and Chen]{trime}
Zhong, Z., Lei, T., and Chen, D.
\newblock Training language models with memory augmentation.
\newblock \emph{CoRR}, abs/2205.12674, 2022.
\newblock \doi{10.48550/arXiv.2205.12674}.
\newblock URL \url{https://doi.org/10.48550/arXiv.2205.12674}.

\end{thebibliography}
\bibliographystyle{icml2023}

\appendix



\section{Complexity Derivation}
\label{section:complexity}

Table \ref{table:complexity_storage} shows the FLOPs per layer for FiD, \memoryname and \modelname. We go into more detail on those values here. For background, we have a source input of length $n_s$, a target output of length $n_t$, and a model with $L$ layers and dimensionality $d$. A linear layer from $k$ to $l$ uses $kl$ FLOPs. 

Each Transformer layer consists of an attention layer, a feedforward layer, and (in case of the decoder) a cross-attention layer. Each attention layer performs query, key, value and output projections, as well as attention score and value computations. The projections map from $d$ to $d$ and use $d^2$ FLOPs per token, totaling $4d^2$ per token. The feedforward layer of a vanilla Transformer maps each token from $d$ to $4d$ and back, consuming $8d^2$ FLOPs per token. 

The attention score and value computation involve taking the inner product of $n_q \cdot n_v$ pairs of vectors of dimension $d$, totaling $n_q n_v d$ FLOPs each. For an FiD encoder layer, the number of queries is equal to the source length, while the number of values is equal to the length of each passage and question (since FiD encodes each passage separately). Therefore the attention score and value computation total $2 n_s n_p d$ FLOPs, leading to total encoder layer FLOPs of
\begin{equation*}
    12 n_s d^2 + 2 n_s n_p d
\end{equation*}
Since retrieved passages are short relative to model dimension ($n_p << d$), especially for larger models, the attention score computation term is very small and we ignore it in our estimate. The derivation for the decoder is similar, except for the presence of cross-attention layers. For cross-attention, the key and value projections apply to the source input and the query and output projections apply to the target output. Therefore, decoder layer FLOPs are given by
\begin{equation*}
    12 n_t d^2 + 2 n_t d^2 + 2 n_s d^2 + 2 n_t n_s d
\end{equation*}
where the attention computation term is again negligible. Totaling FLOPs from both components, we get FLOPs per layer of
\begin{equation*}
    14 n_s d^2 + 14 n_t d^2
\end{equation*}

Finally, \memoryname simply removes the encoder FLOPs, while \modelname multiplies encoder FLOPs by live proportion $\liveprop$, yielding the figures in Table \ref{table:complexity_storage}.
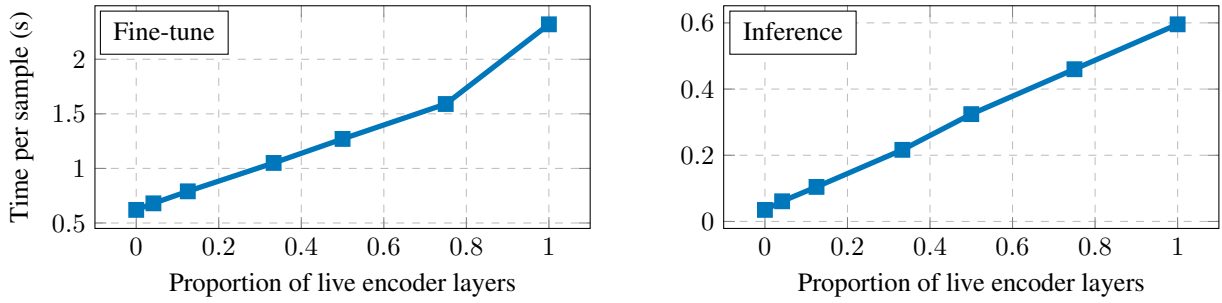
\begin{figure*}
     \centering
     \begin{subfigure}[t]{0.48\textwidth}
        \begin{tikzpicture}[scale=1.0]
            \begin{axis}[
            scale only axis,
            width=0.8\columnwidth,
            height=0.36\columnwidth,
            ylabel={Time per sample (s)},
            xlabel={Proportion of live encoder layers},
            mark=x,
            ymajorgrids=true,
            xmajorgrids=true,
            xminorticks=true,
            grid style=dashed,
            legend columns=1,
            legend cell align=left,
            legend style={
                anchor=south,
                at={(0.14, 0.78)},
            },
        ]
        \addlegendimage{empty legend}\addlegendentry{Fine-tune}
            \addplot[color=xxlcolor,mark=\xxlmark,mark size=2pt,line width=2] table {
                0 0.62  
                0.041666 0.68
                0.125 0.79
                0.333 1.05
                0.5 1.27
                0.75 1.59
            	1.0 2.32
            };             
            \end{axis}
        \end{tikzpicture}
    \end{subfigure}
     \hfill    
    \begin{subfigure}[t]{0.48\textwidth}
        \begin{tikzpicture}[scale=1.0]
            \begin{axis}[
            scale only axis,
            width=0.8\columnwidth,
            height=0.36\columnwidth,
            xlabel={Proportion of live encoder layers},
            mark=x,
            ymajorgrids=true,
            xmajorgrids=true,
            xminorticks=true,
            grid style=dashed,
            legend columns=1,
            legend cell align=left,
            legend style={
                anchor=south,
                at={(0.14, 0.78)},
            },
        ]
                \addlegendimage{empty legend}\addlegendentry{Inference}
            \addplot[color=xxlcolor,mark=\xxlmark,mark size=2pt,line width=2] table {
                0 0.03491967503 
                0.041666 0.06088810956
                0.125 0.1043477324
                0.333 0.2163778237
                0.5 0.3242841923
                0.75 0.4599950808
            	1.0 0.5956714199
            };             
            \end{axis}
        \end{tikzpicture}
    \end{subfigure}
    \caption{Fine-tuning time per sample for \modelname-XXL as a function of live proportion.}
    \label{fig:latency_vs_prop}
\end{figure*}
\begin{figure*}
     \centering
        \begin{subfigure}[t]{0.48\textwidth}
        \begin{tikzpicture}[scale=1.0]
            \begin{axis}[
            scale only axis,
            width=0.80\columnwidth,
            height=0.36\columnwidth,
            ylabel={Exact match},
            xlabel={Time per sample (s)},
            mark=x,
            ymajorgrids=true,
            xmajorgrids=true,
            xminorticks=true,
            grid style=dashed,
            legend columns=1,
            legend cell align=left,
            legend style={
                anchor=south,
                at={(0.85, 0.043)},
            },
        ]
                        \addlegendimage{empty legend}\addlegendentry{Fine-tune}
            \addplot[color=xxlcolor,mark=\xxlmark,mark size=2pt,line width=2] table {
                0.62 42.72  
                0.68 44.26
                0.79 52.89
                1.05 54.95
                1.27 55.0
                1.59 55.14
            	2.32 55.21
            };             
            \end{axis}
        \end{tikzpicture}
    \end{subfigure}
     \hfill        
\begin{subfigure}[t]{0.48\textwidth}
        \begin{tikzpicture}[scale=1.0]
            \begin{axis}[
            scale only axis,
            width=0.80\columnwidth,
            height=0.36\columnwidth,
            xlabel={Time per sample (s)},
            mark=x,
            ymajorgrids=true,
            xmajorgrids=true,
            xminorticks=true,
            grid style=dashed,
            legend columns=1,
            legend cell align=left,
            legend style={
                anchor=south,
                at={(0.85, 0.043)},
            },
        ]
                        \addlegendimage{empty legend}\addlegendentry{Inference}
            \addplot[color=xxlcolor,mark=\xxlmark,mark size=2pt,line width=2] table {
                0.03491967503 42.72  
                0.06088810956 44.26
                0.1043477324 52.89
                0.2163778237 54.95
                0.3242841923 55.0
                0.4599950808 55.14
            	0.5956714199 55.21
            };             
            \end{axis}
        \end{tikzpicture}
    \end{subfigure}        
    \caption{Exact match on Natural Questions as a function of inference time per sample for \modelname-XXL, sweeping over live proportion. Assumes multi-query for inference measurements.}
    \label{fig:perf_vs_latency}
\end{figure*}
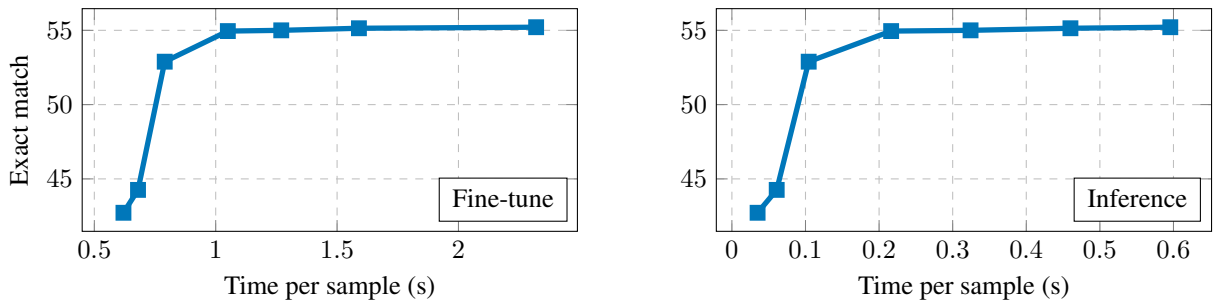
\section{FLOPs vs Latency}
\label{section:flops_vs_latency}

Our main experimental results measure computational cost in terms of FLOPs. However, depending on the model configuration and hardware setting, inference can be slower than FLOPs would suggest due to memory bandwidth constraints. In this section we show that \modelname FLOPs efficiency gains translate to actual throughput improvements.

We consider two settings. First, we investigate how \modelname affects training speed, as memory bandwidth is primarily a bottleneck for autoregressive inference. Second, FiDO~\citep{fido} showed that the memory bandwidth overhead from decoder inference can be drastically reduced with minimal reduction in quality through layer-sparse cross-attention or multi-query attention. At the time of our main experiments we did not have access to T5 checkpoints with multi-query attention, but in production \modelname would be deployed with multi-query attention. Moreover, recent work has shown that multi-query checkpoints can be efficiently obtained from existing multi-head checkpoints~\citep{gqa}. Therefore, we report inference measurements for T5 with multi-query attention.

Figure \ref{fig:latency_vs_prop} shows latency as a function of live proportion, demonstrating that latency varies nearly linearly with the number of remaining live layers. Figure \ref{fig:perf_vs_latency} instead shows the trade-off between latency and quality, yielding the same pattern as Figure \ref{fig:perf_prop_intro}.

\end{document}